\documentclass[10pt,twocolumn,letterpaper]{article}

\usepackage{times}
\usepackage{epsfig}
\usepackage{graphicx}
\usepackage{amsmath}
\usepackage{amssymb}
\usepackage[accsupp]{axessibility}  
\usepackage{subcaption}     
\usepackage{multirow}       
\usepackage{booktabs}       
\usepackage{float}
\usepackage{iccv}
\usepackage[pagebackref=true,breaklinks=true,letterpaper=true,colorlinks,bookmarks=false]{hyperref}

\iccvfinalcopy 

\def\iccvPaperID{8123} 

\ificcvfinal\fi

\begin{document}

\title{Self-Knowledge Distillation with Progressive Refinement of Targets}

\author{%
  Kyungyul Kim$^{1}$\ \ \ 
  ByeongMoon Ji$^{1}$\ \ \ 
  Doyoung Yoon$^{1}$\ \ \ 
  Sangheum Hwang$^{2}$\thanks{Corresponding author}\\
  \\
    $^{1}$LG CNS AI Research, Seoul, South Korea\\
    $^{2}$Seoul National University of Science and Technology, Seoul, South Korea\\
  \texttt{\{kyungyul.kim, jibm, dy0916\}@lgcns.com,
  shwang@seoultech.ac.kr}
}

\maketitle
\ificcvfinal\thispagestyle{empty}\fi

\begin{abstract}
   The generalization capability of deep neural networks has been substantially improved by applying a wide spectrum of regularization methods, e.g., restricting function space, injecting randomness during training, augmenting data, etc. In this work, we propose a simple yet effective regularization method named progressive self-knowledge distillation (PS-KD), which progressively distills a model's own knowledge to soften hard targets (i.e., one-hot vectors) during training. Hence, it can be interpreted within a framework of knowledge distillation as a student becomes a teacher itself. Specifically, targets are adjusted adaptively by combining the ground-truth and past predictions from the model itself. We show that PS-KD provides an effect of hard example mining by rescaling gradients according to difficulty in classifying examples. The proposed method is applicable to any supervised learning tasks with hard targets and can be easily combined with existing regularization methods to further enhance the generalization performance. Furthermore, it is confirmed that PS-KD achieves not only better accuracy, but also provides high quality of confidence estimates in terms of calibration as well as ordinal ranking. Extensive experimental results on three different tasks, image classification, object detection, and machine translation, demonstrate that our method consistently improves the performance of the state-of-the-art baselines. The code is available at \href{https://github.com/lgcnsai/PS-KD-Pytorch}{https://github.com/lgcnsai/PS-KD-Pytorch}.
\end{abstract}

\section{Introduction}

\begin{figure*}[!t]
\begin{center}
    \includegraphics[width=0.8\linewidth]{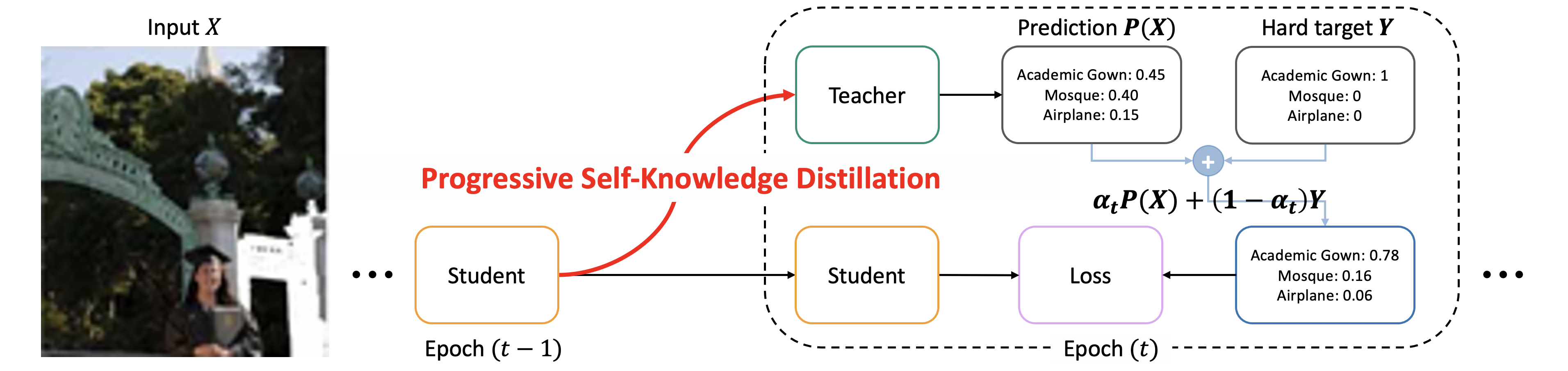}
\end{center}
   \caption{A schematic of PS-KD. At epoch $t$, a student at epoch $(t-1)$ becomes a teacher and a model at epoch $t$ is trained with the soft targets computed as a linear combination of hard targets and the predictions from the teacher.}
\label{fig_scheme}
\end{figure*}

The recent progress made in deep neural networks (DNNs) has significantly improved performance in various tasks related to computer vision as well as natural language processing, e.g., image classification~\cite{ResNet,DenseNet,AlexNet,googlenet}, object detection / segmentation~\cite{maskRCNN,fasterRcnn}, machine translation~\cite{transformer} and language modeling~\cite{languagemodel}. Scaling up of DNN is widely adopted as a promising strategy to achieve higher performances~\cite{pyramidNet,ResNet,efficientNet}. However, deeper networks require a large number of model parameters that need to be learned, which could make models more prone to overfitting. Thus, DNNs typically produce overconfident predictions even for incorrect predictions, and this is because the predictions are highly miscalibrated~\cite{confidence_calibration,high_confidence}.

To improve generalization performance and training efficiency of DNNs, a number of regularization methods have been proposed. The widely employed methods in practice include: $L_1$- and $L_2$-weight decay~\cite{weightdecay1,weightdecay2} to restrict the function space, dropout~\cite{dropout} to inject randomness during training, batch normalization~\cite{batch_nomalisation,howdoes_batchNorm} to accelerate training speed by normalizing internal activations in every layer. 
There also have been several methods that are specifically designed for a particular task. For example, advanced data augmentation techniques that are specific to computer vision tasks such as Cutout~\cite{Cutout}, Mixup~\cite{mixup}, AugMix~\cite{augmix} and CutMix~\cite{CutMix} have shown to boost classification accuracy and also improve robustness and uncertainty of a model. Another effective regularization method is to adjust the targets when they are given in the form of one-hot coded vectors (i.e., hard targets), including label smoothing (LS)~\cite{LabelSmoothing}, label perturbation~\cite{disturblabel}, etc.

Among those methods about adjusting targets, LS~\cite{LabelSmoothing} has been widely applied to many applications~\cite{usels2,transformer,usels1} and has shown to improve generalization performance as well as the quality of confidence estimates (in terms of calibration) on image classification and machine translation tasks~\cite{whendoes}. LS softens a hard target as a smoothed distribution by assigning a small amount of probability mass to non-target classes. However, it is also empirically confirmed that it is not complementary to current advanced regularization techniques. For example, if we utilize LS and CutMix simultaneously for image classification, the performance on both classification and confidence estimation is substantially degraded~\cite{cutmix_empirical}.

One natural question raised on LS could be: is there a more effective strategy to soften hard targets so as to obtain more informative labels? To answer this question, we propose a simple regularization technique named \textit{progressive self-knowledge distillation} (PS-KD) that distills the knowledge in a model itself and uses it for training the model. It means that a student model becomes a teacher model itself, which gradually utilizes its own knowledge for softening hard targets to be more informative during training. Specifically, the model is trained with the soft targets which are computed as a linear combination of the hard targets and the past predictions at a certain epoch, which are adjusted adaptively as training proceeds. 

To justify our proposed method, we have shown that \textit{PS-KD gives more weights to hard-to-learn examples by a gradient rescaling scheme during training}, which clearly reveals that a student model can be enhanced even if trained with a teacher worse than the student (e.g., past predictions). 
The proposed method is easy to implement, can be applied to any supervised learning tasks where the hard targets are given as the ground-truth labels. Moreover, it can be easily combined with current advanced regularization techniques, thereby enhancing further their generalization performance. With this simple method, the generalization ability of DNNs can be greatly improved regarding the target metrics (e.g., accuracy) as well as the quality of confidence estimates in terms of both calibration and ordinal ranking (i.e., misclassification detection)~\cite{AURC,crl}. 

To rigorously evaluate the advantages of PS-KD, we conduct extensive experiments on diverse tasks with popular benchmark datasets: image classification on CIFAR-100 and ImageNet, object detection on PASCAL VOC, and machine translation on IWSLT15 and Multi30k. The experimental results demonstrate that training with PS-KD further enhances the generalization performance of the state-of-the-art baselines. For image classification on CIFAR-100 and ImageNet, our results show that the model trained with PS-KD provides not only better predictive performance, but also high quality of confidence estimates. We further confirm that the advanced image augmentation techniques such as Cutout and CutMix also benefit from PS-KD. From the evaluation on object detection with PASCAL VOC, it is observed that PS-KD makes a model learn better representations compared to the existing approaches. To show the wide applicability of PS-KD, we also conduct the experiments on machine translation, which improve BLEU scores of baselines significantly.

\section{Related Work}

Conventional Knowledge Distillation (KD)~\cite{knowledge_distilling} methods use knowledge from a larger and better performing teacher model to generate soft targets for a student network. Recently, several works have tried to use the student network itself as a teacher, named \textit{self-knowledge distillation} (self-KD)~\cite{SKD_NLP,SKD_distortion,CS-KD}.

One approach to self-KD is to reduce the feature distance between similar inputs from a single network. Xu and Liu~\cite{SKD_distortion} propose a mechanism based on image distortion. Given an image, it generates two separate distorted images by using random mirroring and cropping. Then, model is trained to decrease the distance between features extracted from those two images. Yun~\textit{et al.}~\cite{CS-KD} propose a method called class-wise self-KD (CS-KD) which focuses on distilling knowledge between samples in the same class. For an input $\mathbf{x}$, another data $\mathbf{x}'$ with the same label is randomly sampled, and the KL divergence between predictive distributions from those is minimized during training. 
However, this approach may cause overfitting since it forces all samples in a specific class to have a high density in the learned representation space.

Another approach is to directly use outputs from a teacher whose architecture is exactly the same as a student. This approach is closer to the original notion of KD than the aforementioned approach.
Born-Again Networks (BANs)~\cite{bornagain} first train a network and use this pretrained network as a teacher for next generation. Then, it repeats this process, and thereby, performs multiple generations of KD where the $k$-th generation model is trained with knowledge distilled from the $(k-1)$-th model. Similar to BANs, Yuan~\textit{et al.}~\cite{TF-KD} empirically examine the common belief on KD and suggest the teacher-free KD (TF-KD) method, which uses a pretrained student as a teacher for a single generation. 
Zhang~\textit{et al.}~\cite{SKD_ownteacher} propose a method which divides a network into several blocks and attaches auxiliary classifiers to them independently. These auxiliary classifiers are trained using outputs of the final classifier as a teacher's knowledge. Yang~\textit{et al.}~\cite{snapshot_distil} introduce utilizing a snapshot model at a certain epoch as a teacher: the whole training process is split into multiple mini-generations and the last snapshot model of each mini-generation is used as a teacher for the next mini-generation. These methods require augmenting a network's architecture~\cite{SKD_ownteacher} or several hyperparameters to be carefully tuned, e.g., a learning rate policy and the number of mini-generation~\cite{snapshot_distil}.

Along with previous studies, our proposed PS-KD method uses a model's own predictions as a teacher's knowledge to enhance the generalization performance of DNNs. However, our method provides distinct advantages over them from a practical viewpoint: it does not require a pretraining phase unlike BANs~\cite{bornagain} and TF-KD~\cite{TF-KD}, and can be easily applied to any supervised learning tasks due to its generality and simplicity compared to~\cite{SKD_distortion,SKD_NLP,SKD_ownteacher,snapshot_distil}. More importantly, it is shown that we can enjoy the advantages of KD even if a student is taught by a poor teacher (i.e., lower predictive performance than a student), which is distinguishable from previous works relying on a well-performing teacher.

\section{Self-Knowledge Distillation}

\subsection{Knowledge Distillation as Softening Targets}
KD is a technique to transfer knowledge from one model (i.e., a teacher) to another (i.e., a student), usually from a larger model to a smaller one. The student learns from more informative sources, the predictive probabilities from the teacher, besides one-hot labels. Hence, it can attain a similar performance compared to the teacher although it is usually a much smaller model, and show comparable or even better performance when the student has the same capacity as the teacher~\cite{bornagain, TF-KD}.

For an input $\mathbf{x}$ and a $K$-dimensional one-hot target $\mathbf{y}$, a model produces the logit vector $\mathbf{z}(\mathbf{x})=[z_1(\mathbf{x}),\cdots,z_K(\mathbf{x})]$, and then outputs the predicted probabilities $P(\mathbf{x})=[p_1(\mathbf{x}),\cdots,p_K(\mathbf{x})]$ by a softmax function. Hinton et al.~\cite{knowledge_distilling} suggest to utilize temperature scaling to soften these probabilities for better distillation:
    \begin{equation} 
      \widetilde{p_i}(\mathbf{x};\tau) = {{\exp(z_i(\mathbf{x}) / \tau)} \over {\sum_j \exp(z_j(\mathbf{x}) / \tau)}}
    \end{equation}
where $\tau$ denotes a temperature parameter. By scaling the softmax output $P^{T}(\mathbf{x})$ of the teacher as well as $P^{S}(\mathbf{x})$ of the student, the student is trained with the loss function $\mathcal{L}_{KD}$, given by:
    \begin{equation} \label{eq_kl}
    \begin{split}
      \mathcal{L}_{KD}(\mathbf{x},\mathbf{y}) = & (1-\alpha) H\left(\mathbf{y},P^{S}(\mathbf{x})\right) + \\
      & \alpha\tau^2 H\left(\widetilde{P^T}(\mathbf{x};\tau),\widetilde{P^S}(\mathbf{x};\tau)\right)
    \end{split}
    \end{equation}
where $H$ is a cross-entropy loss and $\alpha$ is a hyperparameter. Note that when the temperature $\tau$ is set to 1, Eq.~(\ref{eq_kl}) is equivalent to the cross-entropy of $P^{S}(\mathbf{x})$ to the soft target, a linear combination of $\mathbf{y}$ and $P^{T}(\mathbf{x})$:
\begin{equation} \label{eq_soft}
  \mathcal{L}_{KD}(\mathbf{x},\mathbf{y}) = H\left( (1-\alpha)\mathbf{y}+\alpha P^{T}(\mathbf{x}),P^{S}(\mathbf{x}) \right) .
\end{equation}
Therefore, the existing methods that use the soft targets for regularization can be interpreted within the framework of knowledge distillation. For example, LS~\cite{whendoes} is equivalent to distilling the knowledge from the teacher which produces uniformly distributed probabilities on any inputs.

\subsection{Distilling Knowledge from the Past Predictions}

We propose a new way of KD, called \textit{progressive self-knowledge distillation} (PS-KD), which distills the knowledge of itself to enhance the generalization capability. In other words, the student becomes the teacher itself, and utilizes its past predictions to have more informative supervisions during training as can be seen in Fig.~\ref{fig_scheme}. Let $P^{S}_{t}(\mathbf{x})$ be the prediction about $\mathbf{x}$ from the student at $t$-th epoch. Then, our objective at $t$-th epoch can be written as:
\begin{equation} \label{eq_past}
  \mathcal{L}_{KD,t}(\mathbf{x},\mathbf{y}) = H\left( (1-\alpha)\mathbf{y}+\alpha P^{S}_{i<t}(\mathbf{x}),P^{S}_{t}(\mathbf{x}) \right).
\end{equation}
Note that using the predictions from $t$-th epoch as the teacher's knowledge is trivial since it will not incur any loss.

The main difference from the conventional KD is that the teacher is not a static model, but dynamically evolves as training proceeds. Among all past models that are candidates for the teacher, we use the model at $(t-1)$-th epoch as the teacher since it can provide most valuable information among the candidates. Concretely, in $t$-th epoch of training, the target for the input $\mathbf{x}$ is softened as $(1-\alpha)\mathbf{y} + \alpha P^{S}_{t-1}(\mathbf{x})$. It is empirically observed that this approach utilizing the past model as a teacher regularizes the model effectively.

One more thing we have to consider is how to determine $\alpha$ in Eq.~(\ref{eq_past}). The $\alpha$ controls how much we are going to trust the knowledge from the teacher. In the conventional KD, the teacher remains unchanged so the $\alpha$ is usually set to a fixed value during training. However, in PS-KD, the reliability of the teacher should be considered since the model generally does not have enough knowledge about data at the early stage of training. To this end, we increase the value of $\alpha$ gradually. Like the learning rate scheduling, there are several strategies to increase the $\alpha$ as a function of the epoch, e.g., step-wise, exponential, linear growth, etc. To minimize the number of hyperparameters involved in the scheduling, we apply the linear growth approach. Therefore, the $\alpha$ at $t$-th epoch is computed as follows:
\begin{equation} \label{eq_alpha}
  \alpha_t = \alpha_T \times {t \over T},
\end{equation}
where $T$ is the total epoch for training and $\alpha_T$ is the $\alpha$ at the last epoch, which is a single hyperparameter to be determined via validation process. Surprisingly, this simple strategy combined with past predictions improves the generalization performance significantly across a wide range of tasks.
To summarize, our objective function at $t$-th epoch can be written as:
\begin{equation} \label{eq_main}
  \mathcal{L}_{KD,t}(\mathbf{x},\mathbf{y}) = H\left( (1-\alpha_t)\mathbf{y}+\alpha_t P^{S}_{t-1}(\mathbf{x}),P^{S}_{t}(\mathbf{x}) \right) .
\end{equation}

\paragraph{Theoretical support.}
We show that \textit{PS-KD pays more attention to hard examples during training} when a hyperparameter $\alpha_t$ is properly set, and \textit{the $\alpha_t$ value should be gradually increased to preserve such an effect of hard example mining}. This effect is realized by example re-weighting, motivated by Proposition 2 in Tang~\textit{et al.}~\cite{understanding_improve_KD}. 
The gradient of $\mathcal{L}_{KD,t}$ in Eq.~\ref{eq_main} with respect to a logit value $z_i$, $i=1,...,K$ for a fixed $\alpha$\footnote{For notational simplicity, we ignore $t$ in $\alpha_t$.} is given by 
\begin{equation} \label{eq-gradient}
  \frac{\partial \mathcal{L}_{KD,t}}{\partial z_i} = \partial^{KD,t}_i = (1-\alpha)(p_{t,i} - y_i) + \alpha (p_{t,i} - p_{t-1,i}).
\end{equation}
Therefore, for $z_{GT}$ where $GT$ denotes the target class,
\begin{equation}
    \begin{split}
        \partial^{KD,t}_{GT} &= (1-\alpha)(p_{t,GT} - 1) + \alpha (p_{t,GT} - p_{t-1,GT}) \\
        &= (p_{t,GT}-1) - \alpha (p_{t-1,GT}-1), \\
    \end{split}
\end{equation}
and for $z_i$, $i\neq{GT}$,
\begin{equation}
    \begin{split}
        \partial^{KD,t}_{i} &= (1-\alpha)p_{t,i} + \alpha (p_{t,i} - p_{t-1,i}) \\
        &= p_{t,i} - \alpha p_{t-1, i}
    \end{split}
\end{equation}
If $\alpha$ is set to be $p_{t,i} - \alpha p_{t-1,i} \geq 0$ for all $i \neq GT$, i.e., $\alpha \leq \min_i \left( p_{t,i}/p_{t-1,i} \right)$, then $(p_{t,GT} - 1) - \alpha (p_{t-1,GT} - 1) < 0$ and $\sum_{i \neq GT} |p_{t,i} - \alpha p_{t-1,i}| = (1 - p_{t,GT}) - \alpha(1 - p_{t-1,GT})$ holds.
Therefore, $L_1$ norm $\sum_{i} |\partial^{KD,t}_i|$ of the gradient can be written as:
\begin{equation} \label{eq-reduced-norm-kd}
\sum_{i} |\partial^{KD,t}_i| = 2(1-p_{t,GT}) - 2 \alpha (1-p_{t-1,GT}) .
\end{equation}

\begin{figure}[t]
  \centering
     \begin{subfigure}[b]{0.23\textwidth}
         \centering
         \includegraphics[width=\textwidth]{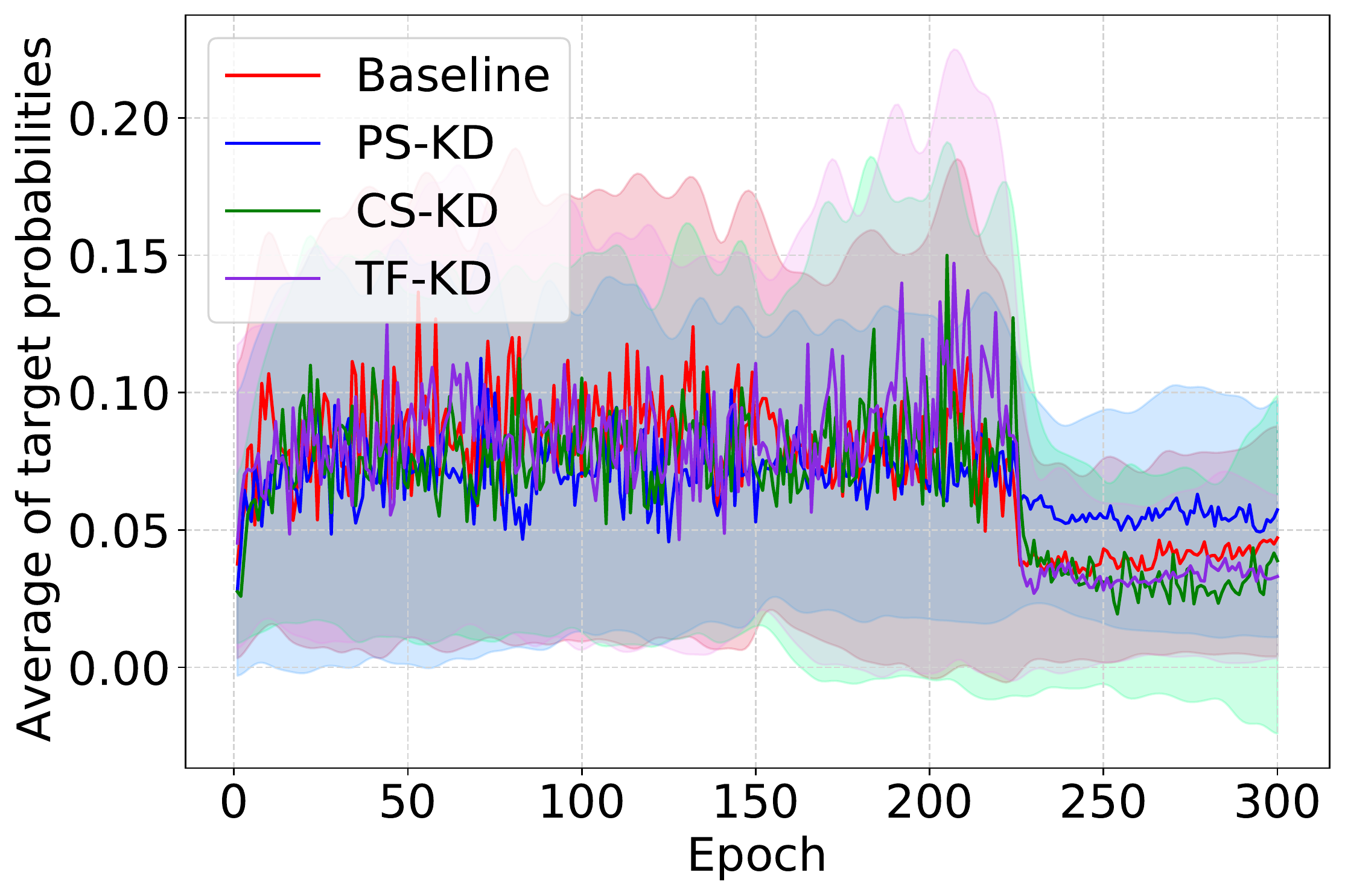}
         \caption{Target probability (for $GT$)}
     \end{subfigure}
     \begin{subfigure}[b]{0.23\textwidth}
         \centering
         \includegraphics[width=\textwidth]{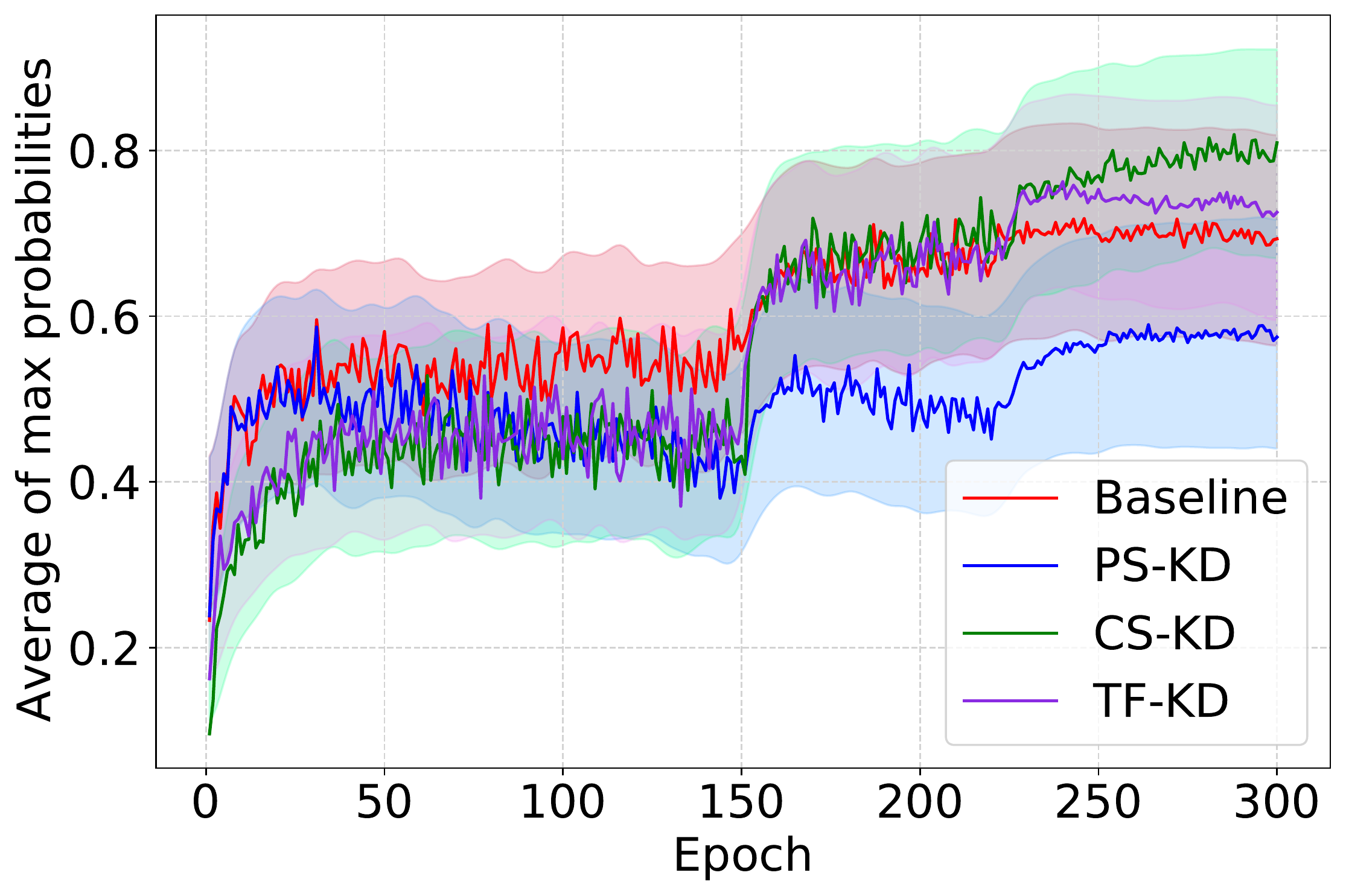}
         \caption{Maximum probability}
     \end{subfigure}
     \vspace{0pt}
     \caption{
        (a) Target and (b) maximum probabilities for 100 hard-to-learn samples (correctly classified less than 50 times during 300 epochs) on CIFAR-100. PS-KD keeps learning from hard examples by giving more weights to them.}
     \label{fig-prob}
     \vspace{-10pt}
\end{figure}

Let us consider the ratio of $L_1$ norms $\sum_{i} |\partial^{KD,t}_i| / \sum_{i} |\partial_i|$, which represents the gradient rescaling factor induced by PS-KD. By combining Eq.~\ref{eq-reduced-norm-kd} and the fact that $\sum_{i} |\partial_i| = 2(1-p_{t,GT})$, the rescaling factor is given by:
\begin{equation}
    \frac{\sum_{i} |\partial^{KD,t}_i|}{ \sum_{i} |\partial_i|} = 
    1- \alpha \left( \frac{1-p_{t-1,GT}}{1-p_{t,GT}} \right)
    \equiv 1-\alpha \left( \frac{\gamma_{t-1}}{\gamma_t} \right) .
\end{equation}
Note that $\gamma$ represents the probability of being incorrect.
Without loss of generality, it can be assumed that $p_{t,GT} \geq p_{t-1,GT}$ and $p_{t,i} \leq p_{t-1,i}$ for all $i \neq GT$ since $P^{S}_{t}$ provides better prediction than $P^{S}_{t-1}$ on training data. 
Therefore, $\gamma_{t-1} \geq \gamma_t$ always holds. A large $\frac{\gamma_{t-1}}{\gamma_t}$ means that predictions on that example are greatly improved during iterations (i.e., easy-to-learn). Conversely, hard-to-learn examples will have a small value.
Consequently, the gradient rescaling factors for hard-to-learn examples are greater than those for easy-to-learn examples, which implies that PS-KD gives more weights on hard-to-learn examples during training, and it is empirically confirmed as shown in Fig.~\ref{fig-prob}. Fig.~\ref{fig-prob} shows that other methods are more overconfident on incorrect predictions than PS-KD, and clearly demonstrates that PS-KD focuses on hard examples implicitly by the gradient rescaling scheme.

To expect such hard example mining effects, $\alpha$ should satisfy the condition described above for all examples. From this, $\alpha$ should be set to a sufficiently small value during an early training phase. As training proceeds, the differences of $\frac{\gamma_{t-1}}{\gamma_t}$'s for all examples become smaller. Therefore, $\alpha$ should be gradually increased to preserve the effect.

\vspace{-10pt}
\paragraph{Implementation.}
For PS-KD, the predictions from the model at $(t-1)$-th epoch are necessary for training at $t$-th epoch. There are two ways to obtain these past predictions. One is to load the model at $(t-1)$-th epoch on memory when $t$-th epoch is started so that the past predictions for softening targets are also computed in forward passes. The other is to save the past predictions on disk in advance during $(t-1)$-th epoch, and read this information to compute the soft targets at $t$-th epoch. These two approaches have pros and cons. The former way may need more GPU memory. On the other hand, the latter way does not need additional GPU memory but requires more space to store past predictions.

The choice of how to obtain the past predictions depends on the task we are dealing with. For example, on machine translation with a large-scale corpus, it is nearly impossible to store the predicted probabilities over all tokens. For this, we can choose the former strategy. Note that softening targets via a moving average~\cite{softtarget} or distilling knowledge with task-specific operations~\cite{SKD_distortion,CS-KD} is not applicable to this task.
In our experiments, we employ an efficient way according to the task, e.g., the past predictions from the model on GPU memory is utilized for the tasks on ImageNet classification and IWSLT15 machine translation.

\section{Experimental Results}
In this section, we show the effectiveness of PS-KD across a variety of tasks including image classification, object detection, and machine translation. More details on datasets, evaluation metrics and experimental settings are available in the supplementary material. All experiments were performed on NVIDIA DGX-1 system with PyTorch~\cite{PyTorch}.

\subsection{CIFAR-100 Classification} \label{section:CIFAR-100 classification}

On CIFAR-100 classification, we consider four CNN models: ResNet~\cite{ResNet}, ResNeXt~\cite{ResNeXt}, DenseNet~\cite{DenseNet}, and PyramidNet~\cite{pyramidNet}. First, we compare PS-KD with LS and recent self-KD methods, CS-KD~\cite{CS-KD} and TF-KD~\cite{TF-KD}, on the architectures we consider. 
Then, we demonstrate that PS-KD is complementary to the existing regularization methods including data augmentation (e.g. Cutout~\cite{Cutout}, CutMix~\cite{CutMix}, etc.), and ensembles.

\paragraph{Experimental settings.} The detailed architectures we consider are PreAct ResNet-18~\cite{preActResNet}, ResNet-101~\cite{ResNet}, ResNeXt-29 (cardinality=8, width=64)~\cite{ResNeXt}, DenseNet-121 (growth rate=32)~\cite{DenseNet} and PyramidNet-200 (widening factor=240)~\cite{pyramidNet}. We follow standard data augmentation schemes: 32$\times$32 random crop after padding with 4 pixels and random horizontal flip.
All CNNs are trained using SGD with a momentum of 0.9 for 300 epochs, and the learning rate is decayed by a factor of 10 at 150 and 225 epochs. For ResNet, ResNeXt, and DenseNet, we set the mini-batch size, a weight decay, and an initial learning rate to 128, 0.0005, and 0.1, respectively. For PyramidNet, the mini-batch size, a weight decay, and an initial learning rate are set to 64, 0.0001, and 0.25, respectively, following~\cite{pyramidNet,CutMix}.

To compare the performance of PS-KD with LS, CS-KD\footnote{\href{https://github.com/alinlab/cs-kd}{https://github.com/alinlab/cs-kd}} and TF-KD\footnote{\href{https://github.com/yuanli2333/Teacher-free-Knowledge-Distillation}{https://github.com/yuanli2333/Teacher-free-Knowledge-Distillation}}, the hyperparameters are set according to those reported in the corresponding studies.
Our PS-KD has only a single hyperparameter $\alpha_T$. To determine the optimal $\alpha_T$, we use randomly sampled 10\% of training data as a validation dataset. In this experiment, we set the $\alpha_T$ to 0.8 which shows the optimal validation performance in terms of accuracy and calibration on ResNet-18.\footnote{The hyperparameters of each method and the ablation study on $\alpha_T$ are provided in the supplementary material~\ref{section:appendix_ablation_effect_alpht_T}.} 
With this parameter value, we then train a model on the entire training dataset for a fair comparison.
Note that the value tuned for ResNet-18 is also used for other all architectures since we expect that PS-KD is fairly robust to the hyperparameter $\alpha_T$, and it is confirmed from the experimental results.

For existing regularization methods to be combined with PS-KD, we also follow the hyperparameter values reported in the literature, for example, the hole size in Cutout is set to 8 and the parameter $\alpha$ of Beta distribution in CutMix is set to 1.
Cutout and CutMix produce randomly synthesized images from two inputs at every iteration. In this case, applying PS-KD with them at the same time is not straightforward. Therefore, for the experiments where PS-KD is combined with Cutout or CutMix, each data selects the regularization method with a probability of 0.5. Simply, PS-KD is applied to the half of the data in a randomly shuffled mini-batch, and Cutout or CutMix is performed on another half of the data.

\begin{table}[t]
\begin{center}
    \resizebox{0.9\linewidth}{!}{%
    \begin{tabular}{l|cc|ccc}
        \toprule
        \begin{tabular}[c]{@{}l@{}}\textbf{Model} \\ \textbf{+ Method}\end{tabular}  & \begin{tabular}[c]{@{}c@{}}\textbf{Top-1} \\ \textbf{Err (\%)}\end{tabular} & \begin{tabular}[c]{@{}c@{}}\textbf{Top-5} \\ \textbf{Err (\%)}\end{tabular} &\textbf{NLL} &\begin{tabular}[c]{@{}c@{}}\textbf{ECE} \\ \textbf{(\%)}\end{tabular}&\begin{tabular}[c]{@{}c@{}}\textbf{AURC} \\ \textbf{($\times10^3$)}\end{tabular}\\
        \midrule
        ResNet-18 &{24.18} & {6.90} & {1.10} & {11.84} & {67.65}\\
        + LS & {20.94} & {6.02} & {0.98} & {10.79} & {57.74}\\
        + CS-KD &{21.30} & {5.70} & {0.88} & {6.24} & {56.56}\\
        + TF-KD &{22.88} & {6.01} & {1.05} & {11.96} & {61.77}\\
        + PS-KD &{\textbf{20.82}} & {\textbf{5.10}} & {\textbf{0.76}} & {\textbf{1.77}} & {\textbf{52.10}}\\
        \midrule
        ResNet-101 &{20.75} & {5.28} & {0.89} & {10.02} & {55.45}\\
        + LS &{19.84} & {5.07} & {0.93} & {\textbf{3.43}} & {95.76}\\
        + CS-KD &{20.76} & {5.62} & {1.02} & {12.18} & {64.44}\\
        + TF-KD &{20.10} & {5.10} & {0.84} & {6.14} & {58.8}\\
        + PS-KD &{\textbf{19.43}} & {\textbf{4.30}} & {\textbf{0.74}} & {6.92} & {\textbf{49.01}}\\
        \midrule
        DenseNet-121 &{20.05} &{4.99} & {0.82} & {7.34} & {52.21}\\
        + LS &{19.80} & {5.46} & {0.92} & {3.76} & {91.06}\\
        + CS-KD &{20.47} & {6.21} & {1.07} & {13.80} & {73.37}\\
        + TF-KD &{19.88} & {5.10} & {0.85} & {7.33} & {69.23}\\
        + PS-KD &{\textbf{18.73}} & {\textbf{3.90}} & {\textbf{0.69}} & {\textbf{3.71}} & {\textbf{45.55}}\\
        \midrule
        ResNeXt-29 &{18.65} & {4.47} & {0.74} & {\textbf{4.17}} & {44.27}\\
        + LS &{17.60} & {4.23} & {1.05} & {22.14} & {41.92}\\
        + CS-KD &{18.26} & {4.37} & {0.80} & {5.95} & {42.11}\\
        + TF-KD &{17.33} & {3.87} & {0.74} & {6.73} & {40.34}\\
        + PS-KD &{\textbf{17.28}}  &{\textbf{3.60}} & {\textbf{0.71}} & {9.15} & {\textbf{39.78}}\\
        \midrule
        PyramidNet &{16.80} &{3.69} & {0.73} & {8.04} & {36.95}\\
        + LS &{17.82} &{4.72} & {0.89} &{3.46} & {105.02}\\
        + CS-KD &{18.31} &{5.70} & {1.17} &{14.70} &{70.05}\\
        + TF-KD &{16.48} & {3.37} & {0.79} & {10.48} & {37.04}\\
        + PS-KD  &{\textbf{15.49}} & {\textbf{3.08}} & {\textbf{0.56}} & {\textbf{1.83}} & {\textbf{32.14}} \\
        \bottomrule
    \end{tabular}}
    \vspace{+5pt}
    \caption{Evaluation results on CIFAR-100 compared to other methods with popular architectures, averaged over three runs. The best result is shown in boldface.}
    \vspace{-10pt}
    \label{table:main result CIFAR-100 result}
\end{center}
\end{table}

\paragraph{Evaluation metrics.} 
We use top-1 and top-5 error as standard performance measures for multi-class classification. We also employ the negative log-likelihood (NLL), expected calibration error (ECE)~\cite{ECE} and the area under the risk-coverage curve (AURC)~\cite{AURC} to evaluate the quality of predictive probabilities in terms of confidence estimation. ECE is a widely used metric to determine whether a model's predictions are well-calibrated, approximating the difference in expectation between classification accuracy and confidence estimates. AURC measures the area under the curve from plotting the risk (i.e., error rate) according to coverage. A low AURC implies that correct and incorrect predictions can be well-separable based on confidence estimates. In these experiments, the maximum class probability is used as a confidence estimator.

\begin{figure*}[t]
  \centering
     \begin{subfigure}[b]{0.18\textwidth}
         \centering
         \includegraphics[width=\textwidth]{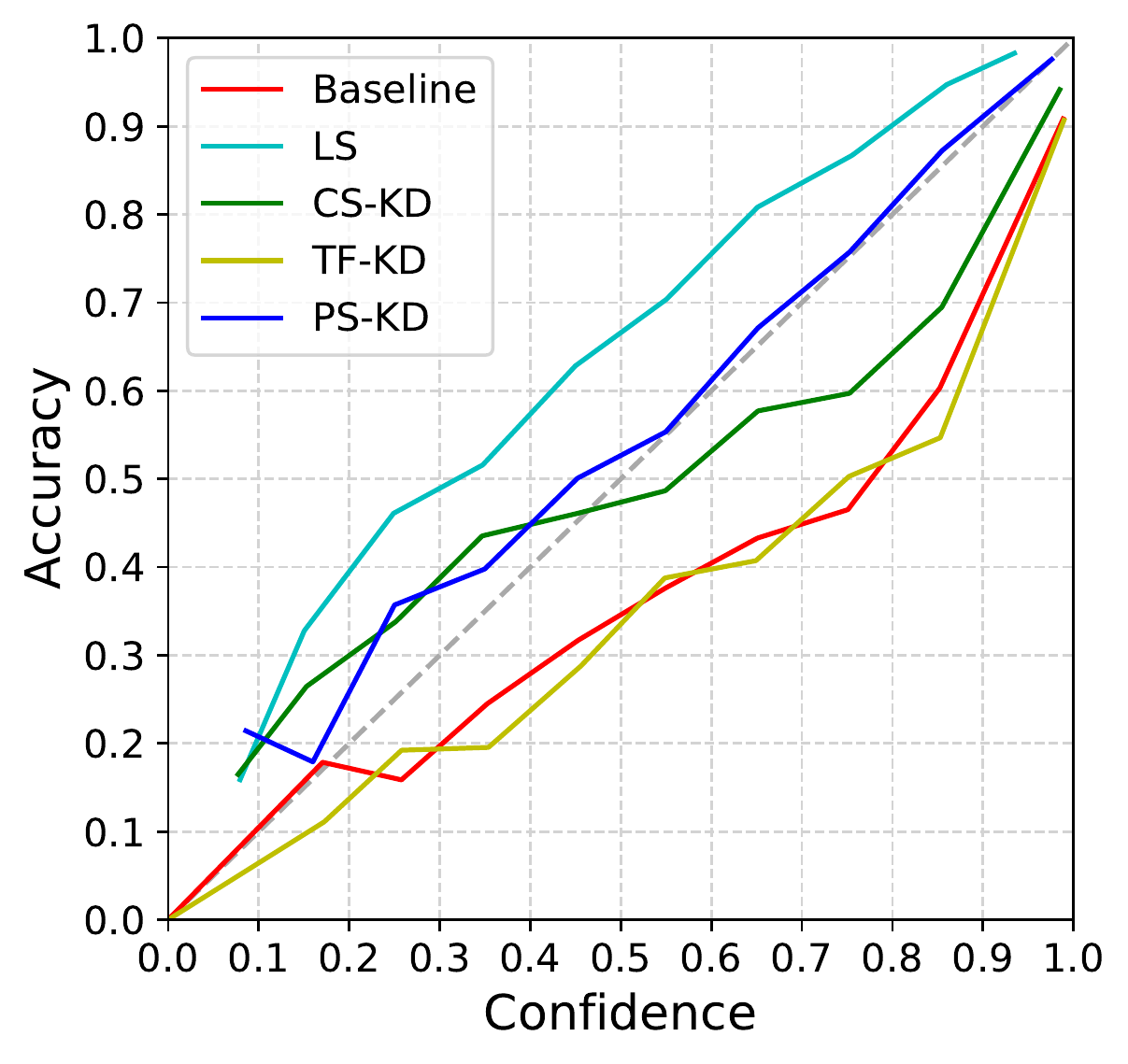}
         \caption{ResNet-18}
         
     \end{subfigure}
     \begin{subfigure}[b]{0.18\textwidth}
         \centering
         \includegraphics[width=\textwidth]{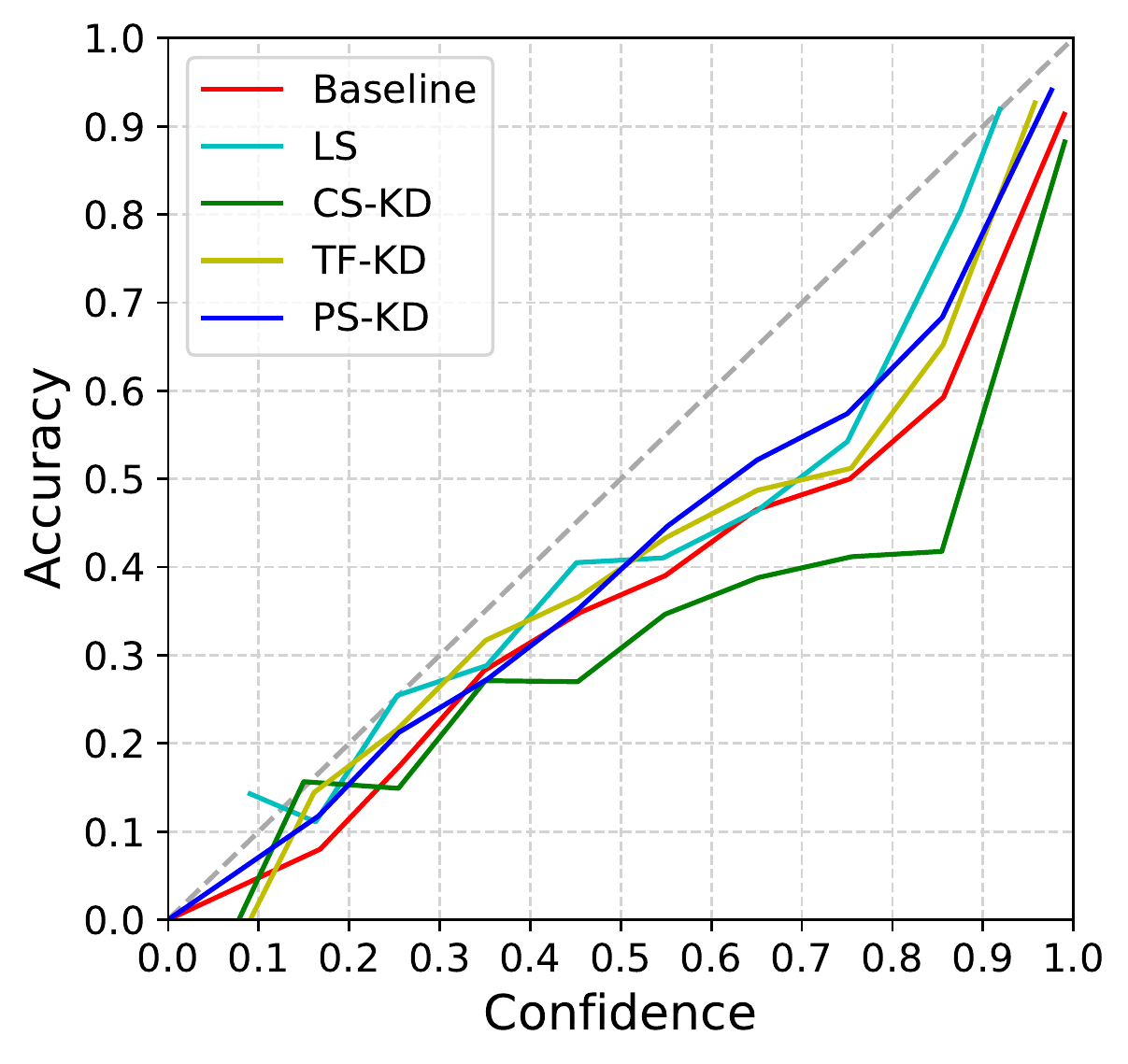}
         \caption{ResNet-101}
         
     \end{subfigure}
     \begin{subfigure}[b]{0.18\textwidth}
         \centering
         \includegraphics[width=\textwidth]{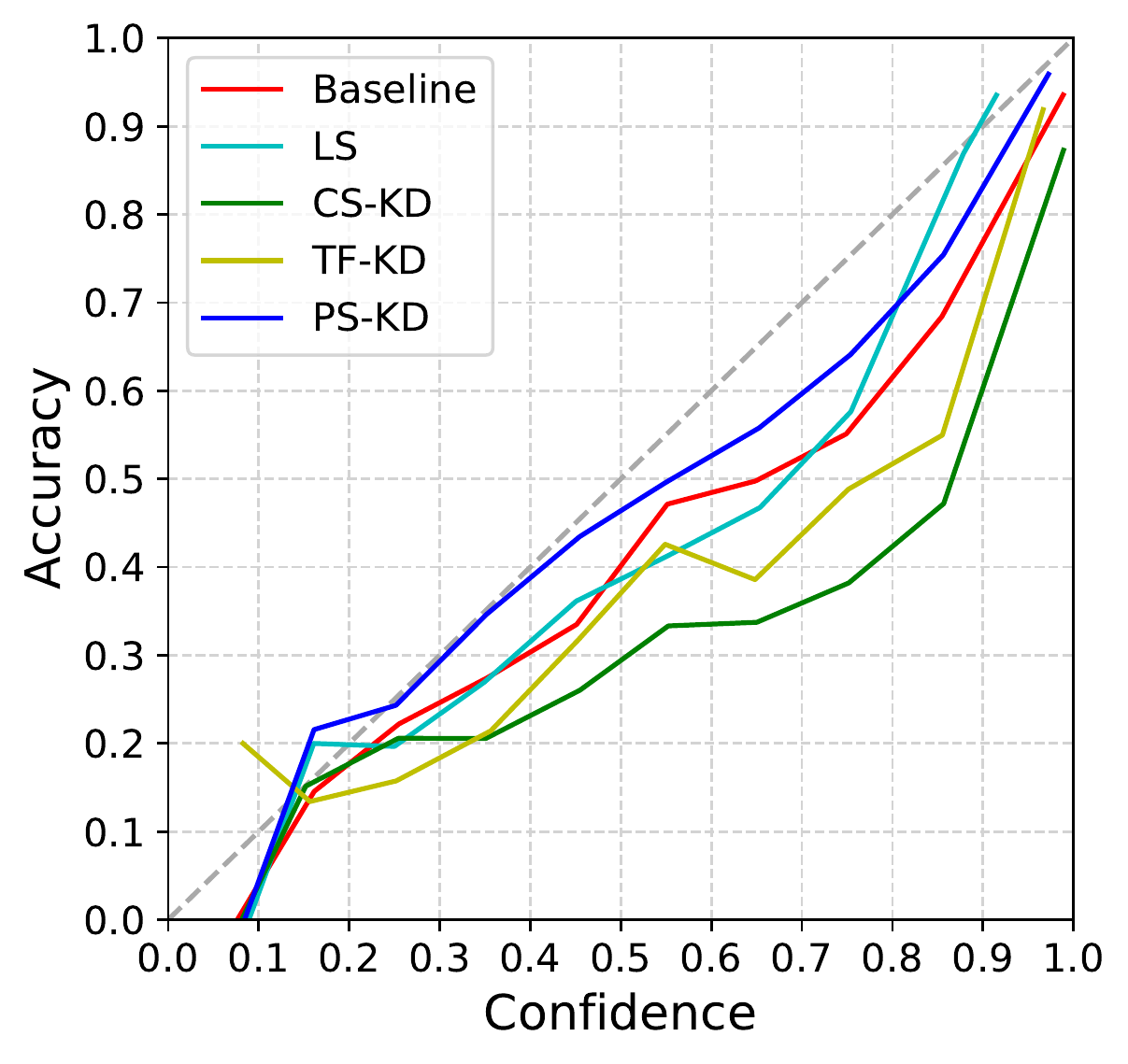}
         \caption{DenseNet-121}
         
     \end{subfigure}
     \begin{subfigure}[b]{0.18\textwidth}
         \centering
         \includegraphics[width=\textwidth]{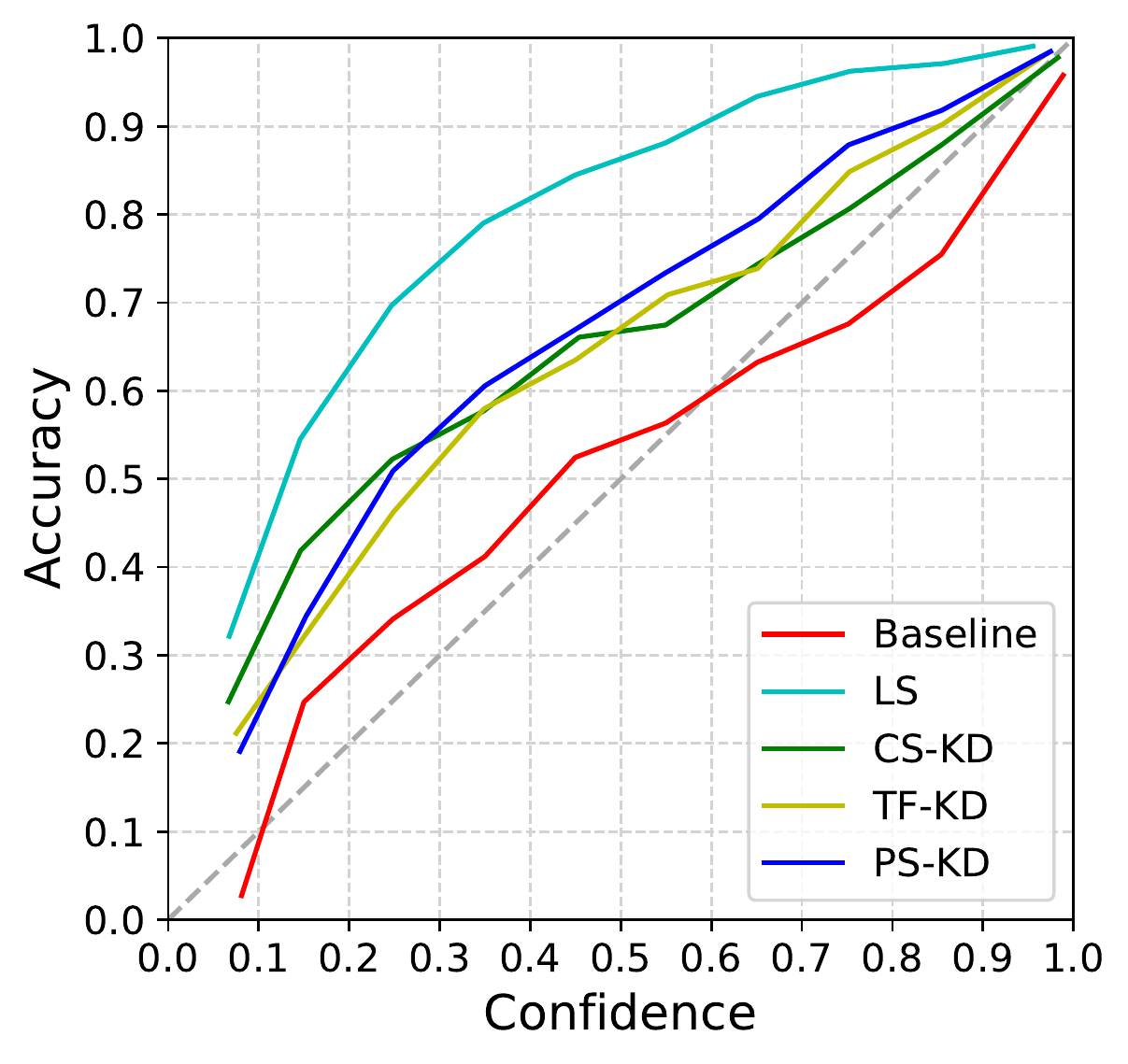}
         \caption{ResNext-29}
         
     \end{subfigure}
     \begin{subfigure}[b]{0.18\textwidth}
         \centering
         \includegraphics[width=\textwidth]{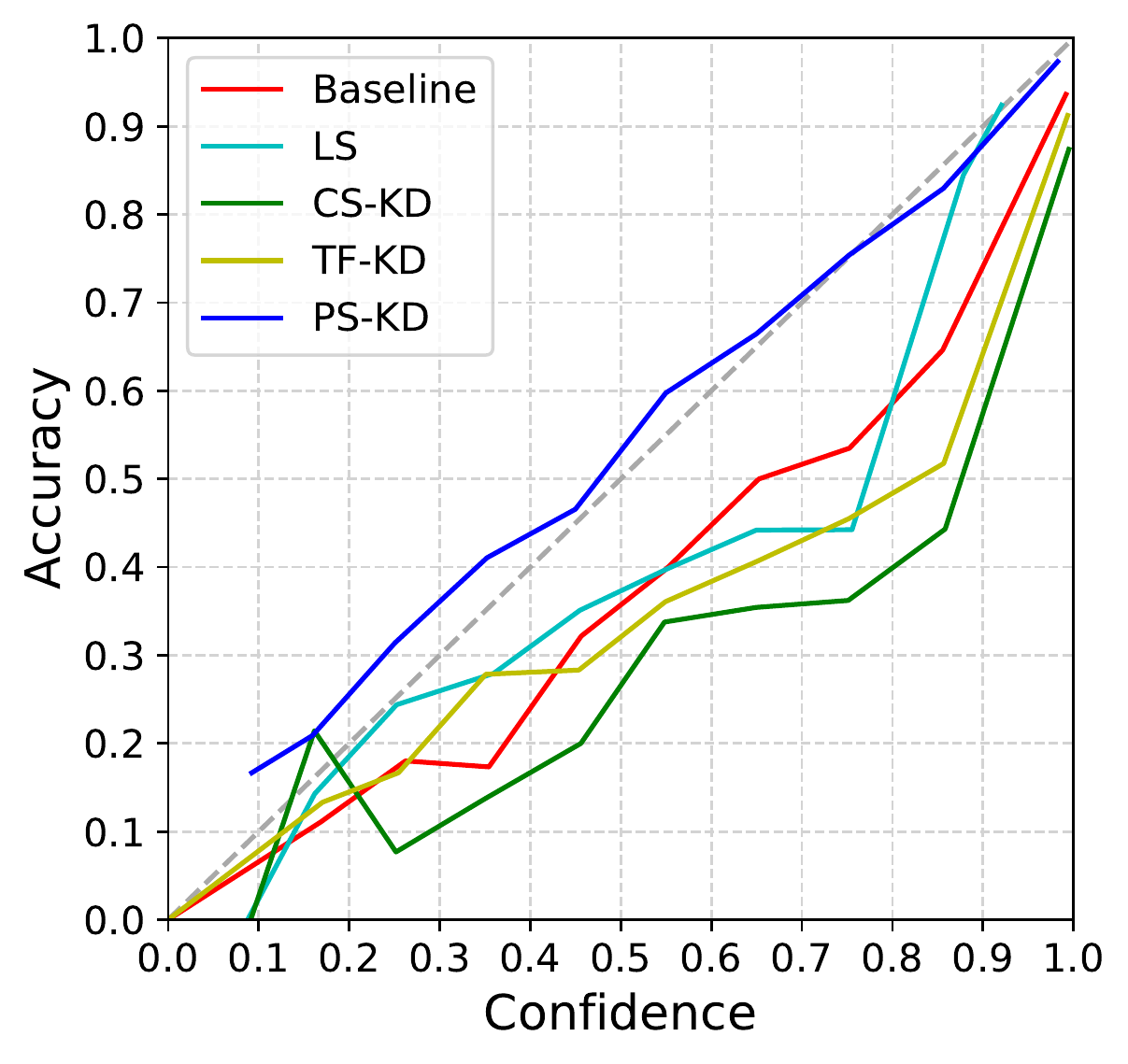}
         \caption{PyramidNet}
     \end{subfigure}
        \vspace{0pt}
        \caption{Calibration plots of all architectures on CIFAR-100. A diagonal dashed line represents perfect calibration.}
        \vspace{-5pt}
        \label{fig:calibration_plot}
\end{figure*}

\paragraph{Result.} 
The comparison results are summarized in Table~\ref{table:main result CIFAR-100 result}. Note that we report average values over three runs. First, we observe that training with PS-KD performs better than baseline and LS in terms of classification accuracy across all architectures, e.g., an improvement of 1.37\% and 0.32\% from baseline and LS on ResNeXt, respectively.
Compared with CS-KD and TF-KD, PS-KD also shows better accuracy while significantly improving the performance on confidence estimation, for example, it improves accuracy by 1.74\% and 1.15\% from CS-KD and TF-KD on DenseNet-121, respectively, and it reduces ECE by 12.87\% and 8.65\% from CS-KD and TF-KD on PyramidNet, respectively. The performance improvement on confidence estimation is consistently observed across all metrics (i.e., NLL, ECE, and AURC) except for the two cases, ECE on ResNet-101 and ResNeXt.\footnote{When $\alpha_T$ is tuned on ResNeXt-29, we observed that most metrics are further improved: for $\alpha_T=0.7$, Top-1=17.06\%, Top-5=3.68\%, NLL=0.69, ECE=6.3\%, and AURC=38.64.} Nevertheless, PS-KD provides robust calibration performance on all architectures, e.g., less than 10\% in ECE, as can be seen from the calibration plots in Fig.~\ref{fig:calibration_plot}.
Interestingly, ResNeXt provides well-calibrated predictions in terms of ECE, and it would be worth investigating which architectural factors of CNNs affect calibration performance. 

From the results, it is confirmed that PS-KD is the only method which shows consistent and robust performance improvement across all metrics.
Importantly, CS-KD performs worse than LS or even baseline in some cases from our experiments. We suspect that these results are caused by the implicit property of CS-KD, which pulls all samples in the same class each other, and it may accelerate overfitting when a model has high capacity or hyperparameters are not properly tuned.

To show that PS-KD can be used in conjunction with other advanced regularization methods, we present the detailed experimental results on PyramidNet in Table~\ref{table:combine regularization CIFAR-100 result}. Compared with Cutout and CutMix on PyramidNet, PS-KD shows slightly higher accuracy while significantly improving performances on confidence estimation.
We observe the top-1 error of 14.82\% when Cutout is combined with PS-KD, which is 1.23\% improvement of Cutout. When PS-KD, CutMix, and SD are utilized simultaneously, the top-1 error from the combination of CutMix and SD is reduced by 0.48\%.
In this setting, it is confirmed again that PS-KD provides a positive effect on confidence estimation: all metrics, NLL, ECE, and AURC, are improved by PS-KD. More experimental results on other self-KD methods are presented in the supplementary material~\ref{section:appendix_combine_result}. These results demonstrate that current state-of-the-art regularization methods benefit from PS-KD in terms of not only classification accuracy, but also confidence estimation. From the previous study~\cite{cutmix_empirical}, it is known that LS might be harmful to generalization performance when applied concurrently with the advanced methods. Our empirical findings reveal that how to soften the hard targets is important and the distilled knowledge from a model itself can be a good source to create more informative targets.

We also examine the ensemble effects of PS-KD. For ensembling, we utilize the three trained models from the previous experiment.
Table~\ref{table:Comparison of Ensemble} shows the performances of ensembles from baseline, CS-KD, TF-KD, and PS-KD models. 
From the results, it is shown that ensembles can also benefit from PS-KD, which implies that PS-KD does not degrade the diversity of independently trained models.

\begin{table}[t]
\begin{center}
  \resizebox{\linewidth}{!}{%
  \begin{tabular}{l|cc|ccc}
    \toprule
    \begin{tabular}[c]{@{}l@{}}\textbf{Model} \\ \textbf{+ Method}\end{tabular} & \begin{tabular}[c]{@{}c@{}}\textbf{Top-1} \\ \textbf{Err (\%)}\end{tabular} & \begin{tabular}[c]{@{}c@{}}\textbf{Top-5} \\ \textbf{Err (\%)}\end{tabular} &\textbf{NLL} &\begin{tabular}[c]{@{}c@{}}\textbf{ECE} \\ \textbf{(\%)}\end{tabular}&\begin{tabular}[c]{@{}c@{}}\textbf{AURC} \\ \textbf{($\times10^3$)}\end{tabular}\\
    \midrule
    PyramidNet & 16.80 & 3.69 & 0.73 & 8.04 & 36.95 \\
    + PS-KD & \textbf{15.49} & \textbf{3.08} & \textbf{0.56} & \textbf{1.83} & \textbf{32.14}\\
    \cmidrule{1-6}
    + Cutout~\cite{Cutout} & 16.05 & 3.42 & 0.67 & 7.15 & 33.20\\
    + Cutout + PS-KD & \textbf{14.82} & \textbf{2.86} & \textbf{0.54} & \textbf{3.69} & \textbf{29.77} \\
    \cmidrule{1-6}
    + CutMix~\cite{CutMix} & 15.62 & 3.38 & 0.68 & 8.16 & 34.60\\
    + CutMix + PS-KD & \textbf{15.03} & \textbf{2.91} & \textbf{0.58} & \textbf{5.81} & \textbf{30.22} \\
    \cmidrule{1-6}
    + CutMix + SD~\cite{Shakedrop} & 14.07 & 2.38 & 0.51 & 3.96 & 28.65 \\
    + CutMix + SD + PS-KD & \textbf{13.59} & \textbf{2.18} & \textbf{0.49} & \textbf{3.46} & \textbf{25.98} \\
    \bottomrule
  \end{tabular}}
  \vspace{+2pt}
  \caption{Performance enhancement of data augmentation methods, Cutout and CutMix, by PS-KD, averaged over three runs. The best result is shown in boldface.}
  \vspace{-15pt}
  \label{table:combine regularization CIFAR-100 result}
\end{center}
\end{table}
\begin{table}[t]
\begin{center}
    \resizebox{0.9\linewidth}{!}{%
    \begin{tabular}{l|cc|ccc}
        \toprule
        \begin{tabular}[c]{@{}l@{}}\textbf{Model} \\ \textbf{+ Method}\end{tabular}  & \begin{tabular}[c]{@{}c@{}}\textbf{Top-1} \\ \textbf{Err (\%)}\end{tabular} & \begin{tabular}[c]{@{}c@{}}\textbf{Top-5} \\ \textbf{Err (\%)}\end{tabular} &\textbf{NLL} &\begin{tabular}[c]{@{}c@{}}\textbf{ECE} \\ \textbf{(\%)}\end{tabular}&\begin{tabular}[c]{@{}c@{}}\textbf{AURC} \\ \textbf{($\times10^3$)}\end{tabular}\\
        \midrule
        ResNet-18 &{21.36}&{5.05} &{0.89} &{4.29} &{54.61} \\
        + CS-KD &{\textbf{18.39}} &{4.21} &{0.74} &{\textbf{3.07}} &{\textbf{43.33}} \\
        + TF-KD &{21.07} &{4.80} &{0.90} &{6.48} &{52.70} \\
        + PS-KD &{18.79} &{\textbf{4.17}} &{\textbf{0.68}} &{5.12} &{44.29} \\
        \midrule
        ResNet-101 &{18.27} &{3.99} &{0.74} &{4.01} &{44.46} \\
        + CS-KD &{17.97} &{4.00} &{0.78} &{4.61} &{43.83} \\
        + TF-KD &{18.16} &{3.97} &{0.71} &{\textbf{1.52}} &{43.66} \\
        + PS-KD &{\textbf{17.26}} &{\textbf{3.42}} &{\textbf{0.63}} &{2.15} &{\textbf{37.85}} \\
        \midrule
        DenseNet-121 &{17.09} &{3.51} &{0.65} &{1.96} &{39.01} \\
        + CS-KD &{17.04} &{3.85} &{0.73} &{4.08} &{42.40} \\
        + TF-KD &{16.96} &{3.32} &{0.71} &{5.36} &{39.31} \\
        + PS-KD &{\textbf{16.25}} &{\textbf{2.85}} &{\textbf{0.57}} &{\textbf{1.92}} &{\textbf{34.64}} \\
        \midrule
        ResNeXt-29 &{16.72} &{3.44} &{\textbf{0.65}} &{\textbf{3.78}} &{36.98} \\
        + CS-KD & {16.86} &{3.51} &{0.75} &{8.58} &{37.06} \\
        + TF-KD&{16.03} &{3.33} &{0.70} &{8.89} &{35.14} \\
        + PS-KD &{\textbf{15.99}} &{\textbf{3.10}} &{0.68} &{11.1} &{\textbf{21.79}} \\
        \midrule
        PyramidNet &{14.58} &{2.85} &{0.60} &{\textbf{2.63}} &{30.04} \\
        + CS-KD & {15.29} &{3.93} &{0.76} &{4.72} &{36.53} \\
        + TF-KD&{14.77} &{2.77} &{0.65} &{5.60} &{30.55} \\
        + PS-KD &{\textbf{14.11}} &{\textbf{2.58}} &{\textbf{0.50}} &{2.79} &{\textbf{27.29}} \\
        \bottomrule
    \end{tabular}}
    \vspace{+5pt}
    \caption{Performance improvement of ensembles by PS-KD. For ensembling, three trained models in Table~\ref{table:main result CIFAR-100 result} are used. The best result is shown in boldface.}
    \label{table:Comparison of Ensemble}
    \vspace{-15pt}
\end{center}
\end{table}

\begin{table}[t]
\begin{center}
    \resizebox{\linewidth}{!}{%
    \begin{tabular}{l|cc|ccc}
        \toprule
        \begin{tabular}[l]{@{}l@{}}\textbf{Model} \\ \textbf{+ Method}\end{tabular}  & \begin{tabular}[c]{@{}c@{}}\textbf{Top-1} \\ \textbf{Err (\%)}\end{tabular} & \begin{tabular}[c]{@{}c@{}}\textbf{Top-5} \\ \textbf{Err (\%)}\end{tabular} &\textbf{NLL} &\begin{tabular}[c]{@{}c@{}}\textbf{ECE} \\ \textbf{(\%)}\end{tabular}&\begin{tabular}[c]{@{}c@{}}\textbf{AURC} \\ \textbf{($\times10^3$)}\end{tabular}\\
        \midrule
        DenseNet-264*~\cite{DenseNet}&{22.15}&{6.12}&{-}&{-}&{-}\\
        ResNeXt-101*~\cite{ResNeXt} &{21.20}&{5.60}&{-}&{-}&{-}\\
        \midrule
        ResNet-152   &{22.19} &{6.19} &{0.88} &{3.84}&{61.79}\\
        + LS         &{21.73} &{5.85} &{0.92} &{3.91}&{68.24}\\
        + CS-KD      &{21.61} &{5.92} &{0.90} &{5.79}&{62.12}\\
        + TF-KD      &{22.76} &{6.43} &{0.91} &{4.70}&{65.28}\\
        + PS-KD      &{21.41} &{5.85} &{0.84} &{2.51}&{61.01}\\
        \cmidrule{1-6}
        + CutMix  &{21.04} & {5.56} &{0.81} &{2.19}&{58.43}\\
        + CutMix + LS &{20.77} &{5.36} &{0.85} &{1.90}&{63.45}\\
        + CutMix + CS-KD &{21.08} &{5.53} &{0.83} &{1.56}&{59.02}\\
        + CutMix + TF-KD &{22.00} &{5.93} &{0.85} &{2.18}&{62.57}\\
        + CutMix + PS-KD &{\textbf{20.76}} & {\textbf{5.34}} &{\textbf{0.80}} & {\textbf{0.54}}&{\textbf{58.25}}\\
        \bottomrule
    \end{tabular}}
    \vspace{+3pt}
    \caption{Top-1/top-5 error, NLL, ECE and AURC results on ImageNet validation dataset.‘*’ denotes results reported in the original papers. The best result is in bold.}
    \label{table:imagenet}
    \vspace{-15pt}
\end{center}
\end{table}

\subsection{ImageNet Classification}\label{section:ImageNet}

In the case of a large-scale dataset like ImageNet~\cite{ImageNet}, the knowledge (i.e., predictions) from the previous snapshot model at $(t-1)$-th epoch might be too outdated since the model at $t$-th epoch further learns from a large number of samples during a single epoch. Nevertheless, we observe that the model benefits from PS-KD even for such a large-scale dataset.
\vspace{-10pt}
\paragraph{Experimental settings.} As a baseline, we train PS-KD using ResNet (depth=152) with standard data augmentation schemes including random resize cropping, random horizontal flip, color jittering, and lighting, following~\cite{CutMix}. We train ResNet-152 for 90 epochs with a weight decay of 0.0001 and an initial learning rate of 0.1, followed by decaying the learning rate by a factor of 10 at 30 and 60 epochs. We employ SGD with a momentum of 0.9 as an optimizer and set the mini-batch size to 256. For LS, we set the hyperparameter $\epsilon$ as 0.1. Since the optimal hyperparameters of CS-KD and TF-KD on ImageNet are not reported in the literature, we conduct a random search over five runs to compare in a fair setting.\footnote{All results can be found in the supplementary material~\ref{section:appendix_RandomSearch}.}
\begin{figure}[t]
    \centering
    \resizebox{\linewidth}{!}{%
    \begin{subfigure}{.35\textwidth}
        \centering
        \includegraphics[width=0.9\linewidth]{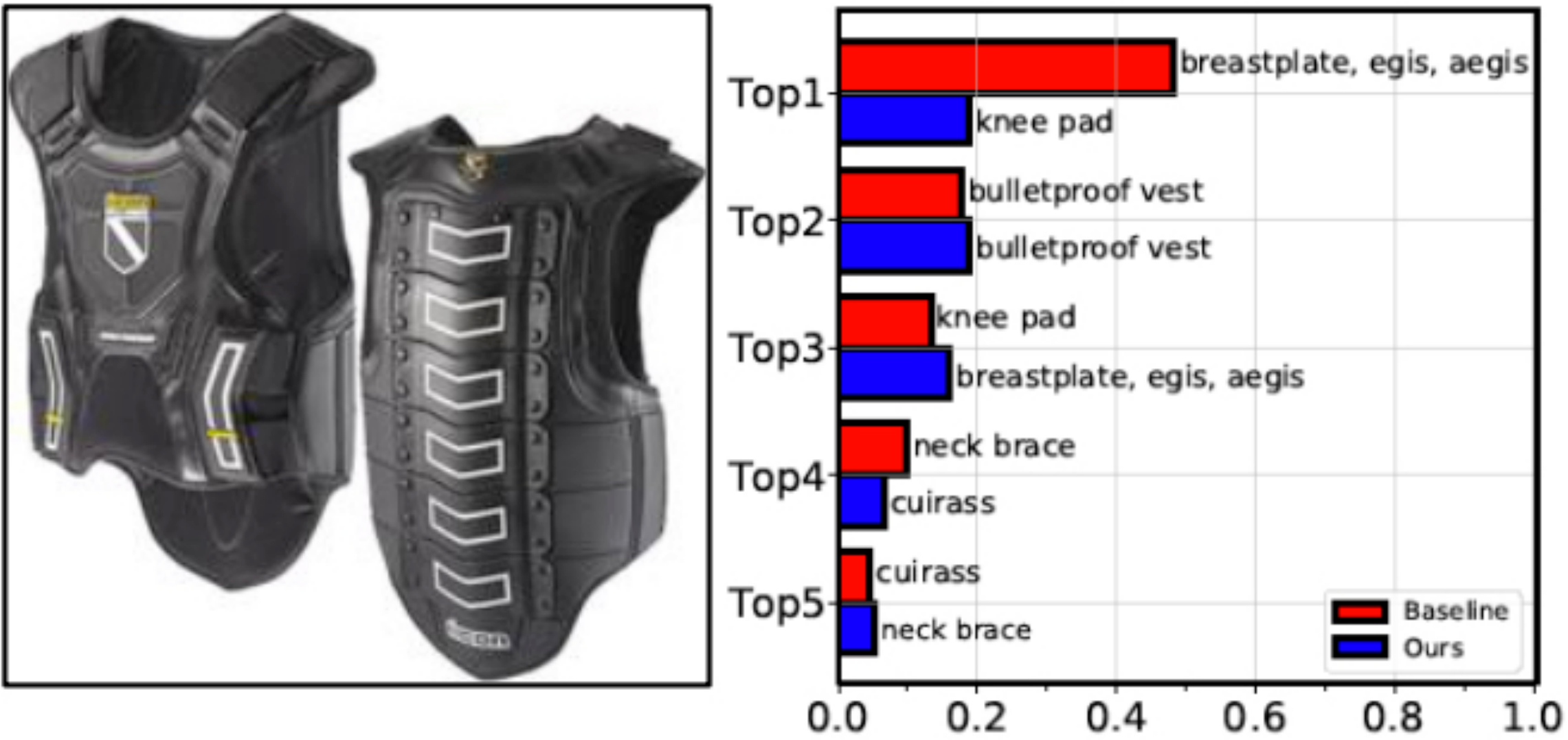}
    \end{subfigure}}
    \resizebox{\linewidth}{!}{%
    \begin{subfigure}{.35\textwidth}
        \centering
        \includegraphics[width=0.9\linewidth]{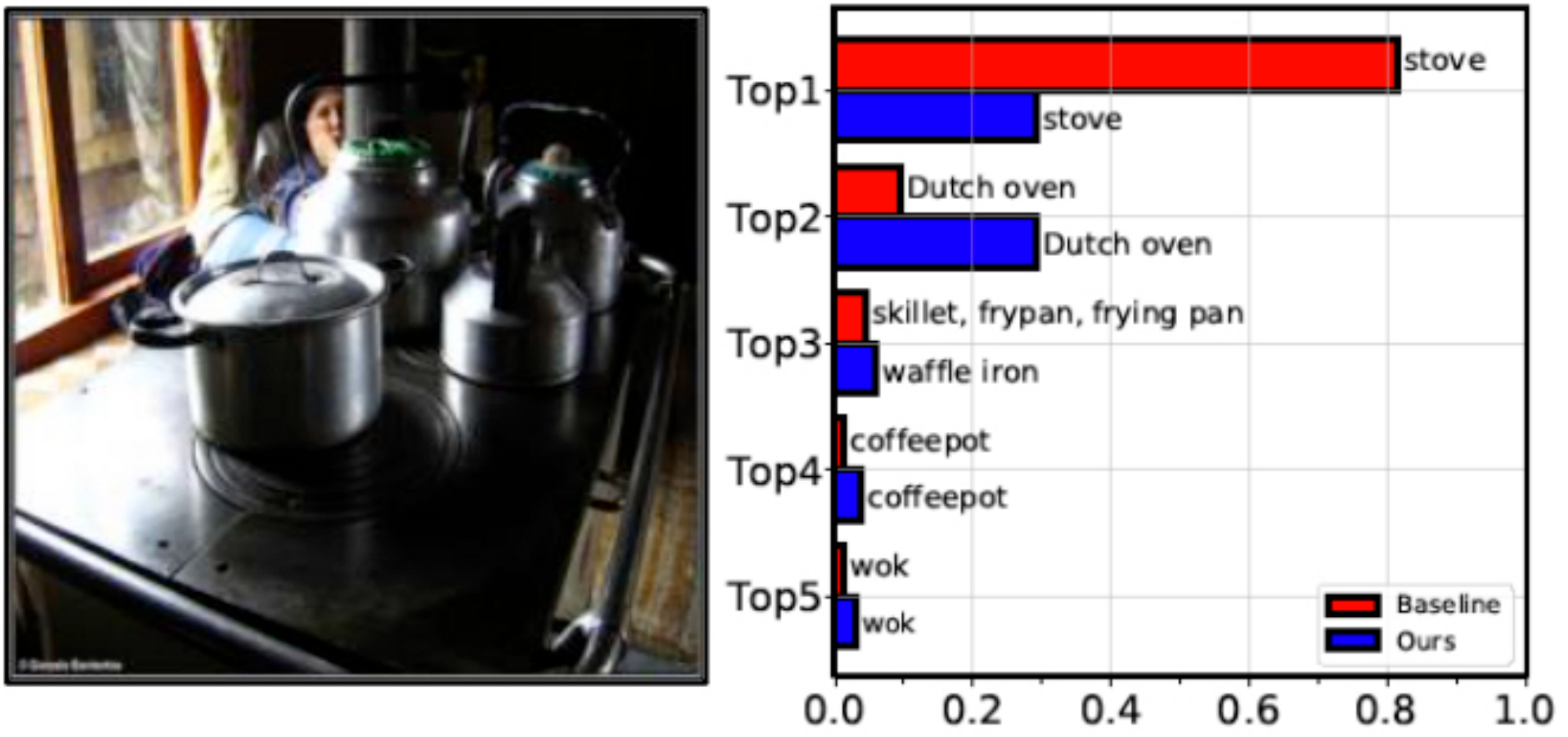}
    \end{subfigure}}
  \vspace{+5pt}
  \caption{Predicted probabilities for samples in the validation dataset from baseline and PS-KD. The ground-truth labels of these images are ``bulletproof vest" (top),  and ``stove" (bottom).}
  \vspace{-10pt}
  \label{figure:imgsmaples}
\end{figure}

\vspace{-10pt}
\paragraph{Result.} Table~\ref{table:imagenet} shows performances evaluated by the metrics used in the previous section. 
Our method shows better accuracy than LS and other self-KD methods, achieving a top-1 error of 21.41\%. Also, PS-KD achieves better performance on confidence estimation, i.e., it reduces ECE by 3.28\% for CS-KD and 2.19\% from TF-KD, respectively. From our validation results provided in the supplementary material~\ref{section:appendix_RandomSearch}, we observe that CS-KD is sensitive to the hyperparameters while PD-KD is much more robust to the hyperparameter $\alpha_T$.
Additionally, PS-KD is further improved on all metrics when combined with CutMix, especially in terms of ECE. It is consistent with the results on CIFAR-100, which demonstrates that PS-KD provides additional benefits to the existing regularization methods. On the other hand, other self-KD methods in conjunction with CutMix are even worse than the vanilla CutMix.
We expect that the performance improvement can be greater if the knowledge from the recent past model is utilized, for example, the predictions from the model at $(t-0.5)$-th epoch. 

Examples of how our PS-KD improves the quality of predicted probability are shown in Fig.~\ref{figure:imgsmaples} (see the supplementary material for more examples). For the top image whose label is ``bulletproof vest", both baseline and PS-KD produce an incorrect prediction. However, PS-KD outputs the class probabilities distributed over the classes that have similar visual characteristics while the baseline outputs overconfident prediction on non-target class. The bottom image is labeled as ``stove" while containing multiple objects including ``coffee pot" and ``stove". Both baseline and PS-KD correctly classify this image, however, PS-KD also produces a high probability on ``Dutch oven" that is visually similar to the objects in the image. These quantitative and qualitative results support the advantage of PS-KD which acts as an effective and strong regularizer.

\subsection{Object Detection}

We also examine that other visual recognition tasks can benefit from PS-KD. For this, we perform the experiment on the task of object detection using PASCAL VOC~\cite{pascalVOC} dataset. We use the 5k VOC 2007 \textit{trainval} and 15k VOC 2012 \textit{trainval} as training sets, and use the PASCAL VOC 2007 \textit{test} as a test set, following~\cite{fasterRcnn,CutMix}. 
As a baseline, Faster R-CNN~\cite{fasterRcnn} is considered, and the improvement of detection performance is examined by replacing the original VGG-16~\cite{VGG} backbone network with a ResNet-152 trained on ImageNet. We utilize six different backbones trained for the previous section: ResNet-152, ResNet-152 with LS, PS-KD, CS-KD, TF-KD, and PS-KD+CutMix. We then fine-tune Faster R-CNN with each backbone network for 10 epochs with a mini-batch size of 1, an initial learning rate of 0.001 decayed by a factor of 10 at 5 epochs. 

\begin{table}[h]
\begin{center}
\small
    \begin{tabular}{l|c}
        \toprule
        \textbf{Method} & {\begin{tabular}[c]{@{}c@{}}\textbf{mAP (IoU $>$ 0.5) (\%)} \end{tabular}}\\
        \midrule
        ResNet-152 &{78.26} \\
        + LS   &{78.44}\\
        + CS-KD &{78.33}\\
        + TF-KD &{78.28}\\
        + PS-KD &{79.50}\\
        + PS-KD + CutMix &{\textbf{79.72}}\\
        \bottomrule
    \end{tabular}
    \vspace{+8pt}
    \caption{Effect of PS-KD as a pretrained backbone network for Faster R-CNN. The mAP value is computed by averaging APs over classes.}
    \label{table:ojbect detection}
    \vspace{-15pt}
\end{center}
\end{table}

As shown in Table~\ref{table:ojbect detection}, ResNet-152 with PS-KD significantly improves the detection performance by 1.06\%, 1.17\%, and 1.22\% of the mean average precision (mAP) compared to ResNet-152 with LS, CS-KD, and TF-KD, respectively.\footnote{APs over all classes are presented in the supplementary material.} Furthermore, PS-KD shows better mAP when it is combined with CutMix. Note that this improvement is achieved by just replacing the backbone network. From this result, it is verified that training with PS-KD provides a strong backbone network, which provides generic representations that can be transferred to other visual recognition tasks. 

\subsection{Machine Translation}
To verify the effectiveness of PS-KD on other tasks rather than multi-class classification, a machine translation task where classification is performed on a token-level, not an input-level is considered.

We use two benchmark datasets including IWSLT15 English to German (EN-DE) and German to English (DE-EN)~\cite{iwslt15}, and Multi30k~\cite{Multi30k} from WMT16~\cite{wmt16}.  
The original purpose of Multi30k is for multimodal learning, consisting of images and descriptions associated with them. For the experiment, we extract only image descriptions written in English and German translations by professional translators. This dataset consists of 29K train data, 1K validation data, and 10K test data with 9,521 vocabularies.

We consider Transformer~\cite{transformer} as our baseline model.\footnote{Experiments were conducted using Fairseq (\href{https://github.com/pytorch/fairseq}{https://github.com/\newline pytorch/fairseq}) toolkit~\cite{fairseq}.} All hyperparameters involved in the architecture and training are set to those reported in~\cite{pay_less_attention}. In specific, we use the architecture with $\textit{N}=6$, $\textit{$d_{model}$}=512$, $\textit{h}=4$, $\textit{$d_{k}$}=64, \textit{$d_{ff}$}=1024$. We train the model for 150 epochs with a maximum of 4,096 tokens per a mini-batch, and employ Adam optimizer~\cite{adam} with $\beta_{1}=0.9$, $\beta_{2} =0.98$.
As a metric, BLEU, commonly used to evaluate the performance on machine translation, is used. The hyperparameter $\alpha_T=0.7$ is determined through validation. 

\begin{table}[h]
\small
    \centering
    \begin{tabular}{llll}
        \toprule
        \multirow{2}{*}{\begin{tabular}[c]{@{}l@{}}\textbf{Model} \\ \textbf{+ Method}\end{tabular}}& \multicolumn{2}{c}{IWSLT15}                                           &\multicolumn{1}{c}{Multi30k}\\
        \cmidrule{2-4}
                              &\multicolumn{1}{c}{EN-DE}           &\multicolumn{1}{c}{DE-EN}          &\multicolumn{1}{c}{DE-EN}   \\
        \midrule
        Transformer  &\multicolumn{1}{c}{28.5} &\multicolumn{1}{c}{34.6} &\multicolumn{1}{c}{29.0}   \\
        + LS         &\multicolumn{1}{c}{29.3} &\multicolumn{1}{c}{35.6} &\multicolumn{1}{c}{29.3}   \\
        + PS-KD      &\multicolumn{1}{c}{\textbf{30.0}} &\multicolumn{1}{c}{\textbf{36.2}} &\multicolumn{1}{c}{\textbf{32.3}}\\
        \bottomrule
    \end{tabular}
    \vspace{+5pt}
    \caption{BLEU scores on Transformer with LS or PS-KD}
    \vspace{-5pt}
    \label{table:translation}
\end{table}

The results are summarized in Table~\ref{table:translation}. Our PS-KD achieves the best BLEU scores on all datasets. Consistent with the results from image classification and object detection, PS-KD shows better performance than the baseline Transformer and that with LS.

\section{Conclusion}
We propose a simple way to improve the generalization performance of DNNs, which distills the knowledge of a model itself to generate more informative targets for training. The targets are softened by using past predictions about data from the model at the previous epoch. We also provide theoretical justification, which shows that our method performs hard example mining implicitly during training. From the experimental results conducted across diverse tasks, we observe that the proposed method is effective to improve the generalization capability of DNNs.

{\small
\bibliographystyle{ieee_fullname}
\bibliography{egbib}

\begin{thebibliography}{10}\itemsep=-1pt

\bibitem{softtarget}
A. {Aghajanyan}.
\newblock Softtarget regularization: An effective technique to reduce
  over-fitting in neural networks.
\newblock In {\em The IEEE International Conference on Cybernetics}, 2017.

\bibitem{iwslt15}
M. Cettolo, J. Niehues, S. Stuker, L. Bentivogli, R. Cattoni, and M. Federico.
\newblock The iwslt 2015 evaluation campaign.
\newblock In {\em International Workshop on Spoken Language Translation}, 2015.

\bibitem{cutmix_empirical}
Sanghyuk Chun, Seong~Joon Oh, Sangdoo Yun, Dongyoon Han, Junsuk Choe, and
  Youngjoon Yoo.
\newblock An empirical evaluation on robustness and uncertainty of
  regularization methods.
\newblock In {\em ICLR Workshop on Uncertainty and Robustness in Deep Leaning},
  2019.

\bibitem{reliability_diagram}
M. Degroot and S. Fienberg.
\newblock The comparison and evaluation of forecasters.
\newblock {\em The Statistician}, 32:12--22, 1983.

\bibitem{ImageNet}
J. Deng, W. Dong, R. Socher, L.-J. Li, K. Li, and L. Fei-Fei.
\newblock {ImageNet: A Large-Scale Hierarchical Image Database}.
\newblock In {\em The IEEE Conference on Computer Vision and Pattern
  Recognition}, 2009.

\bibitem{Cutout}
Terrance Devries and Graham~W. Taylor.
\newblock Improved regularization of convolutional neural networks with cutout.
\newblock {\em arXiv preprint arXiv:1708.04552}, 2017.

\bibitem{Multi30k}
Desmond Elliott, Stella Frank, Khalil Sima'an, and Lucia Specia.
\newblock Multi30k: Multilingual english-german image descriptions.
\newblock In {\em ACL Workshop on Vision and Language}, 2016.

\bibitem{pascalVOC}
Mark Everingham, Luc Van~Gool, Christopher~KI Williams, John Winn, and Andrew
  Zisserman.
\newblock The pascal visual object classes (voc) challenge.
\newblock {\em International Journal of Computer Vision}, 88(2):303--338, 2010.

\bibitem{bornagain}
Tommaso Furlanello, Zachary~C. Lipton, Michael Tschannen, Laurent Itti, and
  Anima Anandkumar.
\newblock Born again neural networks.
\newblock In {\em International Conference on Machine Learning}, 2018.

\bibitem{AURC}
Yonatan Geifman, Guy Uziel, and Ran El{-}Yaniv.
\newblock Bias-reduced uncertainty estimation for deep neural classifiers.
\newblock In {\em International Conference on Learning Representations}, 2019.

\bibitem{confidence_calibration}
Chuan Guo, Geoff Pleiss, Yu Sun, and Kilian~Q. Weinberger.
\newblock On calibration of modern neural networks.
\newblock In {\em International Conference on Machine Learning}, 2017.

\bibitem{SKD_NLP}
Sangchul Hahn and Heeyoul Choi.
\newblock Self-knowledge distillation in natural language processing.
\newblock {\em arXiv preprint arXiv:1908.01851}, 2019.

\bibitem{pyramidNet}
Dongyoon Han, Jiwhan Kim, and Junmo Kim.
\newblock Deep pyramidal residual networks.
\newblock In {\em The IEEE Conference on Computer Vision and Pattern
  Recognition}, 2017.

\bibitem{maskRCNN}
K. {He}, G. {Gkioxari}, P. {Dollár}, and R. {Girshick}.
\newblock Mask r-cnn.
\newblock In {\em The IEEE International Conference on Computer Vision}, 2017.

\bibitem{ResNet}
Kaiming He, Xiangyu Zhang, Shaoqing Ren, and Jian Sun.
\newblock Deep residual learning for image recognition.
\newblock In {\em The IEEE Conference on Computer Vision and Pattern
  Recognition}, 2016.

\bibitem{preActResNet}
Kaiming He, Xiangyu Zhang, Shaoqing Ren, and Jian Sun.
\newblock Identity mappings in deep residual networks.
\newblock In {\em European Conference on Computer Vision}, 2016.

\bibitem{augmix}
Dan Hendrycks, Norman Mu, Ekin~Dogus Cubuk, Barret Zoph, Justin Gilmer, and
  Balaji Lakshminarayanan.
\newblock Augmix: A simple method to improve robustness and uncertainty under
  data shift.
\newblock In {\em International Conference on Learning Representations}, 2020.

\bibitem{knowledge_distilling}
Geoffrey Hinton, Oriol Vinyals, and Jeffrey Dean.
\newblock Distilling the knowledge in a neural network.
\newblock In {\em NIPS Workshop on Deep Learning and Representation Learning},
  2015.

\bibitem{DenseNet}
Gao Huang, Zhuang Liu, Laurens van~der Maaten, and Kilian~Q. Weinberger.
\newblock Densely connected convolutional networks.
\newblock In {\em The IEEE Conference on Computer Vision and Pattern
  Recognition}, 2017.

\bibitem{batch_nomalisation}
Sergey Ioffe and Christian Szegedy.
\newblock Batch normalization: Accelerating deep network training by reducing
  internal covariate shift.
\newblock In {\em International Conference on Learning Representations}, 2015.

\bibitem{languagemodel}
Rafal J{\'{o}}zefowicz, Oriol Vinyals, Mike Schuster, Noam Shazeer, and Yonghui
  Wu.
\newblock Exploring the limits of language modeling.
\newblock {\em arXiv preprint arXiv:1602.02410}, 2016.

\bibitem{adam}
Diederik~P Kingma and Jimmy Ba.
\newblock Adam: A method for stochastic optimization.
\newblock In {\em International Conference on Learning Representations}, 2015.

\bibitem{AlexNet}
Alex Krizhevsky, Ilya Sutskever, and Geoffrey~E Hinton.
\newblock Imagenet classification with deep convolutional neural networks.
\newblock In {\em Advances in Neural Information Processing Systems}, 2012.

\bibitem{weightdecay1}
Anders Krogh and John~A. Hertz.
\newblock A simple weight decay can improve generalization.
\newblock In {\em Advances in Neural Information Processing Systems}, 1992.

\bibitem{crl}
Jooyoung Moon, Jihyo Kim, Younghak Shin, and Sangheum Hwang.
\newblock Confidence-aware learning for deep neural networks.
\newblock In {\em International Conference on Machine Learning}, 2020.

\bibitem{whendoes}
Rafael M{\"u}ller, Simon Kornblith, and Geoffrey~E Hinton.
\newblock When does label smoothing help?
\newblock In {\em Advances in Neural Information Processing Systems}, 2019.

\bibitem{high_confidence}
Anh~Mai Nguyen, Jason Yosinski, and Jeff Clune.
\newblock Deep neural networks are easily fooled: High confidence predictions
  for unrecognizable images.
\newblock In {\em The IEEE Conference on Computer Vision and Pattern
  Recognition}, 2015.

\bibitem{weightdecay2}
Steven~J Nowlan and Geoffrey~E Hinton.
\newblock Simplifying neural networks by soft weight-sharing.
\newblock {\em Neural Computation}, 4(4):473--493, 1992.

\bibitem{fairseq}
Myle Ott, Sergey Edunov, Alexei Baevski, Angela Fan, Sam Gross, Nathan Ng,
  David Grangier, and Michael Auli.
\newblock fairseq: A fast, extensible toolkit for sequence modeling.
\newblock In {\em North American Chapter of the Association for Computational
  Linguistics}, 2019.

\bibitem{ECE}
Mahdi Pakdaman~Naeini, Gregory Cooper, and Milos Hauskrecht.
\newblock Obtaining well calibrated probabilities using bayesian binning.
\newblock In {\em AAAI Conference on Artificial Intelligence}, 2015.

\bibitem{PyTorch}
Adam Paszke, Sam Gross, Francisco Massa, Adam Lerer, James Bradbury, Gregory
  Chanan, Trevor Killeen, Zeming Lin, Natalia Gimelshein, Luca Antiga, Alban
  Desmaison, Andreas Kopf, Edward Yang, Zachary DeVito, Martin Raison, Alykhan
  Tejani, Sasank Chilamkurthy, Benoit Steiner, Lu Fang, Junjie Bai, and Soumith
  Chintala.
\newblock Py{T}orch: An imperative style, high-performance deep learning
  library.
\newblock In {\em Advances in Neural Information Processing Systems}, 2019.

\bibitem{usels2}
Esteban Real, Alok Aggarwal, Yanping Huang, and Quoc~V. Le.
\newblock Regularized evolution for image classifier architecture search.
\newblock In {\em AAAI Conference on Artificial Intelligence}, 2019.

\bibitem{fasterRcnn}
Shaoqing Ren, Kaiming He, Ross Girshick, and Jian Sun.
\newblock Faster r-cnn: Towards real-time object detection with region proposal
  networks.
\newblock In {\em Advances in Neural Information Processing Systems}, 2015.

\bibitem{howdoes_batchNorm}
Shibani Santurkar, Dimitris Tsipras, Andrew Ilyas, and Aleksander Madry.
\newblock How does batch normalization help optimization?
\newblock In {\em Advances in Neural Information Processing Systems}, 2018.

\bibitem{VGG}
Karen Simonyan and Andrew Zisserman.
\newblock Very deep convolutional networks for large-scale image recognition.
\newblock In {\em International Conference on Learning Representations}, 2015.

\bibitem{wmt16}
Lucia Specia, Stella Frank, Khalil Sima{'}an, and Desmond Elliott.
\newblock A shared task on multimodal machine translation and crosslingual
  image description.
\newblock In {\em Association for Computational Linguistics}, 2016.

\bibitem{dropout}
Nitish Srivastava, Geoffrey Hinton, Alex Krizhevsky, Ilya Sutskever, and Ruslan
  Salakhutdinov.
\newblock Dropout: A simple way to prevent neural networks from overfitting.
\newblock {\em Journal of Machine Learning Research}, 15:1929--1958, 2014.

\bibitem{googlenet}
Christian Szegedy, Wei Liu, Yangqing Jia, Pierre Sermanet, Scott Reed, Dragomir
  Anguelov, Dumitru Erhan, Vincent Vanhoucke, and Andrew Rabinovich.
\newblock Going deeper with convolutions.
\newblock In {\em The IEEE Conference on Computer Vision and Pattern
  Recognition}, 2015.

\bibitem{LabelSmoothing}
Christian Szegedy, Vincent Vanhoucke, Sergey Ioffe, Jon Shlens, and Zbigniew
  Wojna.
\newblock Rethinking the inception architecture for computer vision.
\newblock In {\em The IEEE Conference on Computer Vision and Pattern
  Recognition}, 2016.

\bibitem{efficientNet}
Mingxing Tan and Quoc~V. Le.
\newblock Efficient{N}et: Rethinking model scaling for convolutional neural
  networks.
\newblock In {\em International Conference on Machine Learning}, 2019.

\bibitem{understanding_improve_KD}
Jiaxi Tang, Rakesh Shivanna, Zhe Zhao, Dong Lin, Anima Singh, Ed~H. Chi, and
  Sagar Jain.
\newblock Understanding and improving knowledge distillation.
\newblock {\em arXiv preprint arXiv:2002.03532}, 2021.

\bibitem{transformer}
Ashish Vaswani, Noam Shazeer, Niki Parmar, Jakob Uszkoreit, Llion Jones,
  Aidan~N Gomez, Lukasz Kaiser, and Illia Polosukhin.
\newblock Attention is all you need.
\newblock In {\em Advances in Neural Information Processing Systems}, 2017.

\bibitem{pay_less_attention}
Felix Wu, Angela Fan, Alexei Baevski, Yann~N. Dauphin, and Michael Auli.
\newblock Pay less attention with lightweight and dynamic convolutions.
\newblock In {\em International Conference on Learning Representations}, 2019.

\bibitem{disturblabel}
Lingxi Xie, Jingdong Wang, Zhen Wei, Meng Wang, and Qi Tian.
\newblock Disturblabel: Regularizing {CNN} on the loss layer.
\newblock {\em CoRR}, abs/1605.00055, 2016.

\bibitem{ResNeXt}
Saining Xie, Ross Girshick, Piotr Dollar, Zhuowen Tu, and Kaiming He.
\newblock Aggregated residual transformations for deep neural networks.
\newblock In {\em The IEEE Conference on Computer Vision and Pattern
  Recognition}, 2017.

\bibitem{SKD_distortion}
Ting{-}Bing Xu and Cheng{-}Lin Liu.
\newblock Data-distortion guided self-distillation for deep neural networks.
\newblock In {\em AAAI Conference on Machine Learning}, 2019.

\bibitem{Shakedrop}
Y. {Yamada}, M. {Iwamura}, T. {Akiba}, and K. {Kise}.
\newblock Shakedrop regularization for deep residual learning.
\newblock {\em IEEE Access}, 7:186126--186136, 2019.

\bibitem{snapshot_distil}
Chenglin Yang, Lingxi Xie, Chi Su, and Alan~L. Yuille.
\newblock Snapshot distillation: Teacher-student optimization in one
  generation.
\newblock In {\em Proceedings of the IEEE/CVF Conference on Computer Vision and
  Pattern Recognition}, 2019.

\bibitem{TF-KD}
Li Yuan, Francis~EH Tay, Guilin Li, Tao Wang, and Jiashi Feng.
\newblock Revisiting knowledge distillation via label smoothing regularization.
\newblock In {\em Proceedings of the IEEE/CVF Conference on Computer Vision and
  Pattern Recognition}, pages 3903--3911, 2020.

\bibitem{CutMix}
Sangdoo Yun, Dongyoon Han, Seong~Joon Oh, Sanghyuk Chun, Junsuk Choe, and
  Youngjoon Yoo.
\newblock Cutmix: Regularization strategy to train strong classifiers with
  localizable features.
\newblock In {\em The IEEE International Conference on Computer Vision}, 2019.

\bibitem{CS-KD}
Sukmin Yun, Jongjin Park, Kimin Lee, and Jinwoo Shin.
\newblock Regularizing class-wise predictions via self-knowledge distillation.
\newblock In {\em The IEEE Conference on Computer Vision and Pattern
  Recognition}, 2020.

\bibitem{mixup}
Hongyi Zhang, Moustapha Ciss{\'{e}}, Yann~N. Dauphin, and David Lopez{-}Paz.
\newblock mixup: Beyond empirical risk minimization.
\newblock In {\em International Conference on Learning Representations}, 2018.

\bibitem{SKD_ownteacher}
L. {Zhang}, J. {Song}, A. {Gao}, J. {Chen}, C. {Bao}, and K. {Ma}.
\newblock Be your own teacher: Improve the performance of convolutional neural
  networks via self distillation.
\newblock In {\em The IEEE International Conference on Computer Vision}, 2019.

\bibitem{usels1}
Barret Zoph, Vijay Vasudevan, Jonathon Shlens, and Quoc~V. Le.
\newblock Learning transferable architectures for scalable image recognition.
\newblock In {\em The IEEE Conference on Computer Vision and Pattern
  Recognition}, 2017.

\end{thebibliography}
}

\newpage    
\onecolumn

\appendix
\label{sec:supplemental}
\renewcommand\thefigure{S\arabic{figure}}
\renewcommand\thetable{S\arabic{table}}

\section{Image Classification}
\subsection{Evaluation Metrics}
\paragraph{ECE.} Expected calibration error (ECE)~\cite{ECE} is a widely used metric for evaluating confidence calibration performance. To estimate the expected gap between accuracy and confidence, it partitions samples into total $M$ bins, $B_m$ for  $m=1,...,M$, by confidence. Then, each bin $B_m$ contains samples with confidence within $[{{m - 1} \over M},{m \over M}]$. With this binning, ECE is defined as follows,
\begin{equation*}
    ECE = {1 \over n} \sum\limits^M_{m=1} |B_m| \times |\text{Acc}(B_m) - \text{Conf}(B_m)|
\end{equation*}
where $n$ is number of samples, $\text{Acc}(B_m)$ represents accuracy of samples in $B_m$, and $\text{Conf}(B_m)$ represents average confidence of samples in $B_m$. The lower value of ECE indicates that a model is well-calibrated.

The reliability diagram~\cite{reliability_diagram,confidence_calibration} and calibration plot are visualization tools to show how well confidence of a model is calibrated by plotting accuracy against confidence values.

\paragraph{AURC.} Area under risk-coverage curve (AURC)~\cite{AURC} measures how well predictions are ordered by confidence values. Given a classifier, we can define a selective classifier with a threshold which covers only samples with higher confidence than the threshold. Then, coverage can be defined as the proportion of covered samples (i.e., not rejected samples by the selective classifier) to the entire dataset. Risk is defined as an error rate computed by using the covered samples. Therefore, as coverage increases from 0 to 1, risk approaches to the top-1 error on the entire data. AURC is defined as the area under the risk-coverage curve. If a model has a low AURC value, it means that correct and incorrect predictions from the model are well-separable by confidence values.
\subsection{Methods}
\paragraph{Label smoothing.} Szegedy et al.~\cite{LabelSmoothing} proposes a method named label smoothing which improves the performance of deep learning models by adjusting one-hot targets to be soft targets. Soft targets $\mathbf{y}_{LS}$ are computed as a weighted sum of the hard targets $\mathbf{y}$ and the uniform distribution over classes, i.e.,
\begin{equation*}
    \mathbf{y}_{LS} = (1 - \epsilon)\mathbf{y} + {\epsilon \over K}
\end{equation*}
where $\epsilon$ is a smoothing parameter and $K$ is the number of classes.

\paragraph{Cutout.} Cutout~\cite{Cutout} is a simple regularization method designed for image classification. Motivated by dropout and image augmentation, Cutout generates a partially occluded version of input samples, which can be interpreted as an augmented data by applying the structured dropout to an input space. In detail, a square-shaped region with the predefined size is randomly selected on an input image, and that region is zeroed-out during training.

\paragraph{CutMix.} Yun et al.~\cite{CutMix} suggests a method inspired by Cutout~\cite{Cutout} and Mixup~\cite{mixup}. This method generates a new training sample $(\Tilde{x}, \Tilde{y})$ from two samples $(x_a, y_a)$ and $(x_b, y_b)$. From $x_a$, a rectangular region with bounding box coordinates $(r_x, r_y, r_w, r_h)$ will be sampled as a patch. Then, the region of the same coordinates in $x_b$ will be replaced by the patch to generate $\Tilde{x}$. 
For the generated sample $\Tilde{x}$, its target $\Tilde{y}$ is defined as
\begin{equation*}
    \Tilde{y} = \lambda y_a + (1 - \lambda) y_b
\end{equation*}
where $\lambda$ denotes the combination ratio sampled from the uniform distribution $(0,1)$.

\paragraph{ShakeDrop.} ShakeDrop~\cite{Shakedrop} is a regularization technique designed for ResNet and its variants. This method gives regularization effect by replacing residual blocks to ShakeDrop blocks. Let an input be $x$ and an output of residual block be $F(x)$, then the output of $l$-th ShakeDrop block $G(x)$ is defined as,
\begin{equation*}
    G(x) = \begin{cases}
        x + (b_l + \alpha - b_l\alpha) F(x), \; ~~\text{for the train-forward phase} \\
        x + (b_l + \beta - b_l\beta) F(x), \; ~~\text{for the train-backward phase} \\
        x + E[b_l + \beta - b_l\beta] F(x), \; \text{for test phase} \\
    \end{cases}
\end{equation*}
where $\alpha$, $\beta$ are independent uniform random variables and $b_l$ is a Bernoulli random variable with probability $P(b_l=1)=p_l$, which is a parameter with a linear decay according to the block index $l$:
\begin{equation*}
    p_l = 1 - {l \over L}(1-P_L)
\end{equation*}
where $L$ is the total number of building blocks and $P_L$ is an initial parameter. In our experiments, we use $P_L=0.5$ as suggested in~\cite{Shakedrop}.

\subsection{Datasets}
CIFAR-100 is a dataset for multi-class image classification. It consists of 50K training images and 10K test images of 32${\times}$32 resolutions with 100 classes, and has the same number of images per class. The ImageNet is a large-scale dataset. It consists of 1.2M training images and 50K validation images of various resolutions with 1K classes. It contains some images that have multiple objects. In training, we use an input image that is resized to 256${\times}$256, and it is randomly cropped to have a size of 224${\times}$224. For inference, we resize an image as 256${\times}$256 and perform the center crop to have a 224${\times}$224 sized input.

\subsection{Experimental Results on CIFAR-100}
\paragraph{Hyperparameters.}
For LS, we use the smoothing parameter $\epsilon$ of 0.1. For CS-KD\footnote{CS-KD implementation:\href{https://github.com/alinlab/cs-kd}{https://github.com/alinlab/cs-kd}}, we set the temperature $\tau$ to 4, and the weight $\lambda_{cls}$ to 1~\cite{CS-KD}. For TF-KD\footnote{TF-KD implementation: \href{https://github.com/yuanli2333/Teacher-free-Knowledge-Distillation}{https://github.com/yuanli2333/Teacher-free-Knowledge-Distillation}}, we use TF-KD$_{self}$ method presented in~\cite{TF-KD}. The hyperparameters, the temperature $\tau$ and weight $\alpha$, for ResNet-18, DenseNet-121 and ResNeXt-29 are set to the values reported in~\cite{TF-KD}. For ResNet-101 and PyramidNet, we use the temperature $\tau=20$ and weight $\alpha=0.95$, which are most widely used settings in the paper.

\begin{figure}[H]
  \vspace{2pt}
  \begin{center}
  \includegraphics[height=6cm]{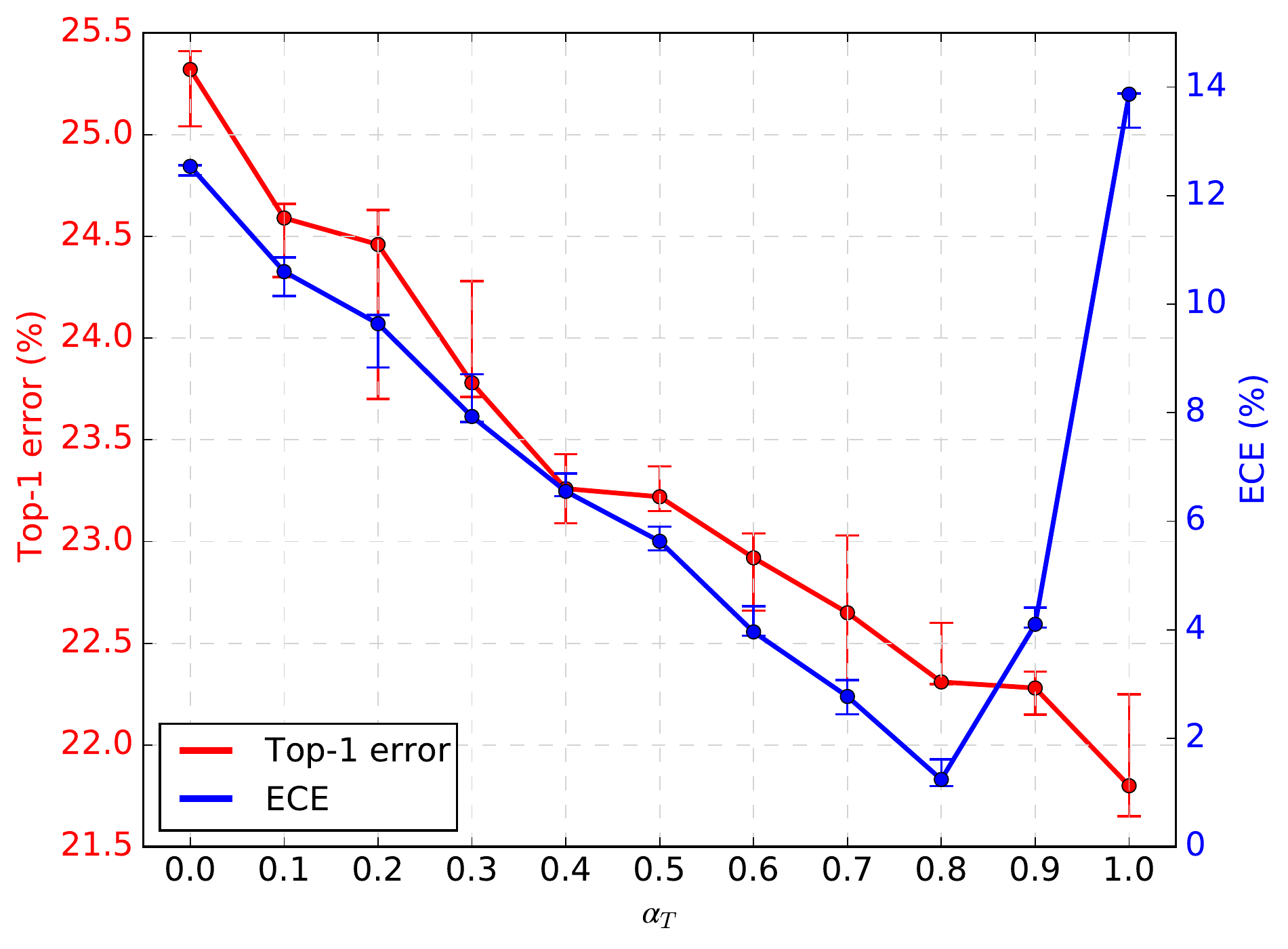}
  \end{center}
  \vspace{-15pt}
  \caption{Validation top-1 error and ECE according to $\alpha_T$ from three repeated experiments on CIFAR-100 for ResNet-18. $\alpha_T=0.8$  is chosen as the best one and used for all other experiments.}
  \label{figure:ablation}
\end{figure}

\begin{figure}[H]
  \centering
     \begin{subfigure}[b]{0.33\textwidth}
         \includegraphics[width=\textwidth]{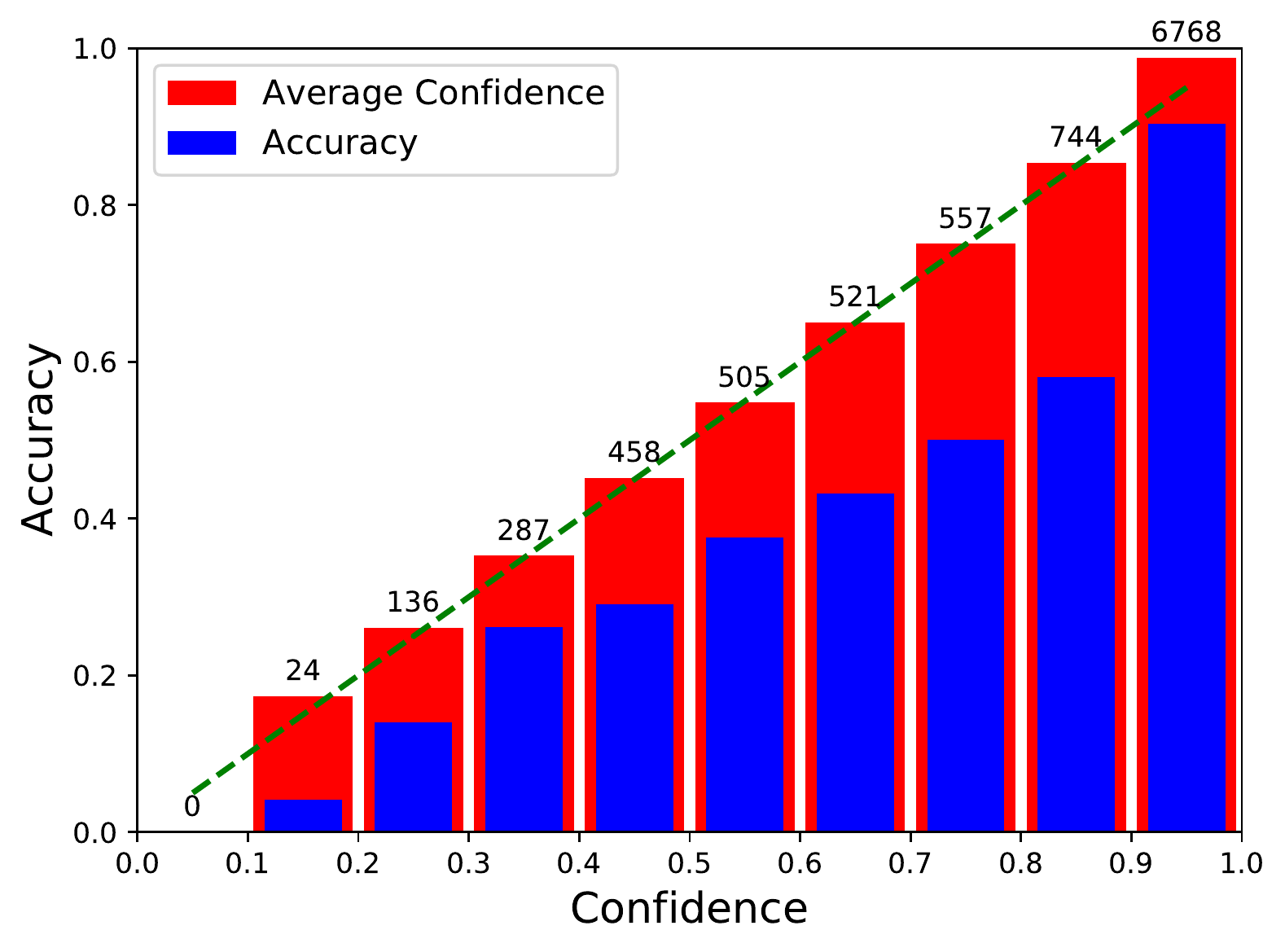}
         \caption{$\alpha_T=0.0$}
     \end{subfigure}
     \begin{subfigure}[b]{0.33\textwidth}
         \includegraphics[width=\textwidth]{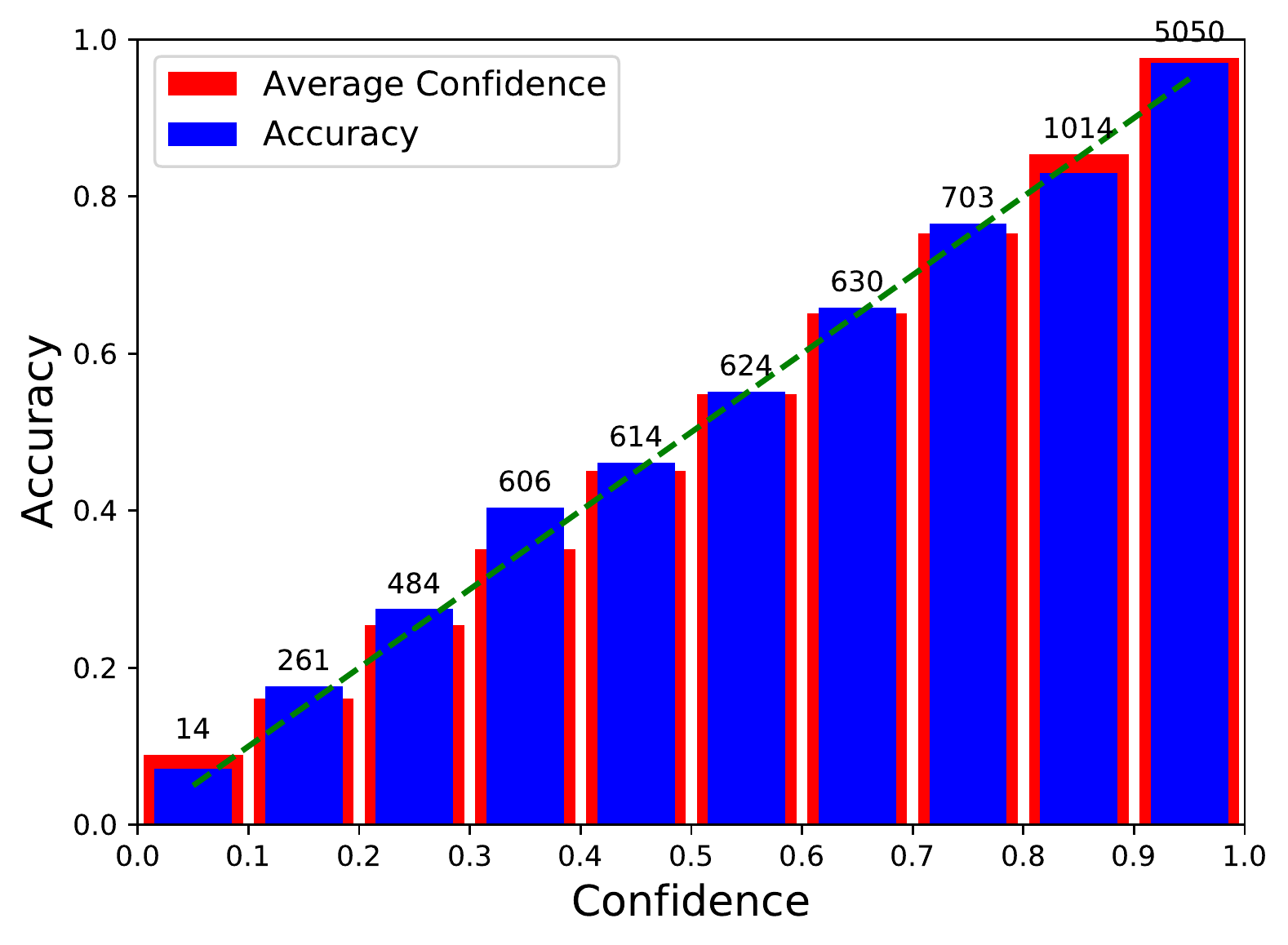}
         \caption{$\alpha_T=0.8$}
     \end{subfigure}
     \begin{subfigure}[b]{0.33\textwidth}
         \includegraphics[width=\textwidth]{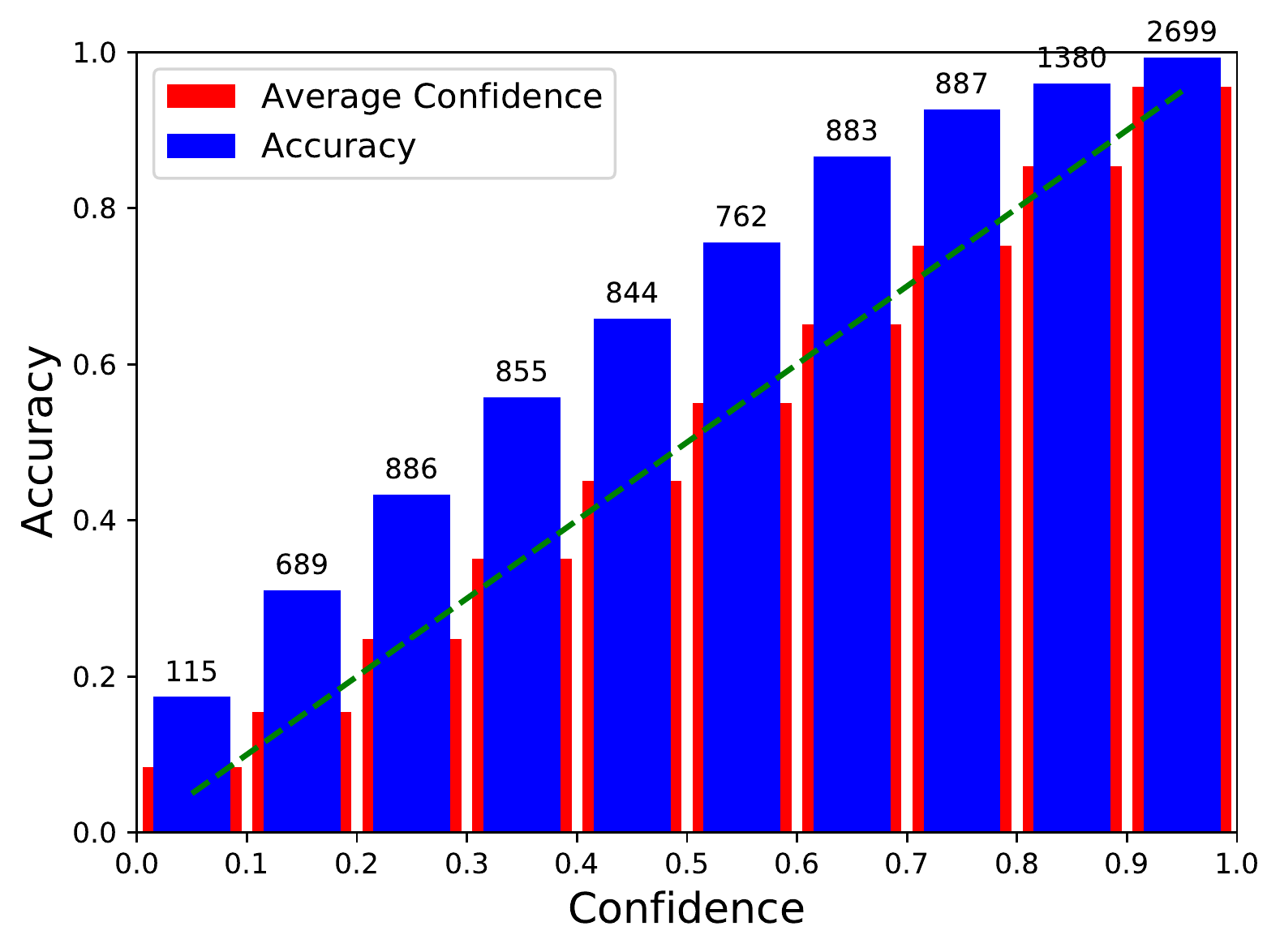}
         \caption{$\alpha_T=1.0$}
     \end{subfigure}
        \vspace{-10pt}
        \caption{Reliability diagrams on the validation dataset of CIFAR100 with ResNet-18+PS-KD. The number on the top of each bin represents the number of samples belonging to that bin. }
        \label{fig:supple_reliability_of_ablation}
\end{figure}

\paragraph{Ablation study on the hyperparameter $\alpha_T$ of PS-KD.}
\label{section:appendix_ablation_effect_alpht_T}
To investigate the effect of our hyperparameter $\alpha_T$, we provide the validation performances in terms of top-1 error and ECE on CIFAR-100 with ResNet-18. 
The results are given in Fig.~\ref{figure:ablation}. Considering both top-1 error and ECE metrics, we determine the optimal $\alpha_T$ as 0.8.
For $\alpha_T > 0.8$, we observe that ECE suffers from PS-KD while top-1 accuracy still improves, implying that PS-KD with a large value of $\alpha_T > 0.8$ tends to produce underconfident predictions as can be seen in Fig.~\ref{fig:supple_reliability_of_ablation}.
Fig.~\ref{fig:supple_reliability_of_ablation} shows the reliability diagrams on the validation dataset with PS-KD. PS-KD with $\alpha_T = 0.8$ shows best calibration performance.

\begin{figure}[H]
  
  \centering
  \includegraphics[height=7cm]{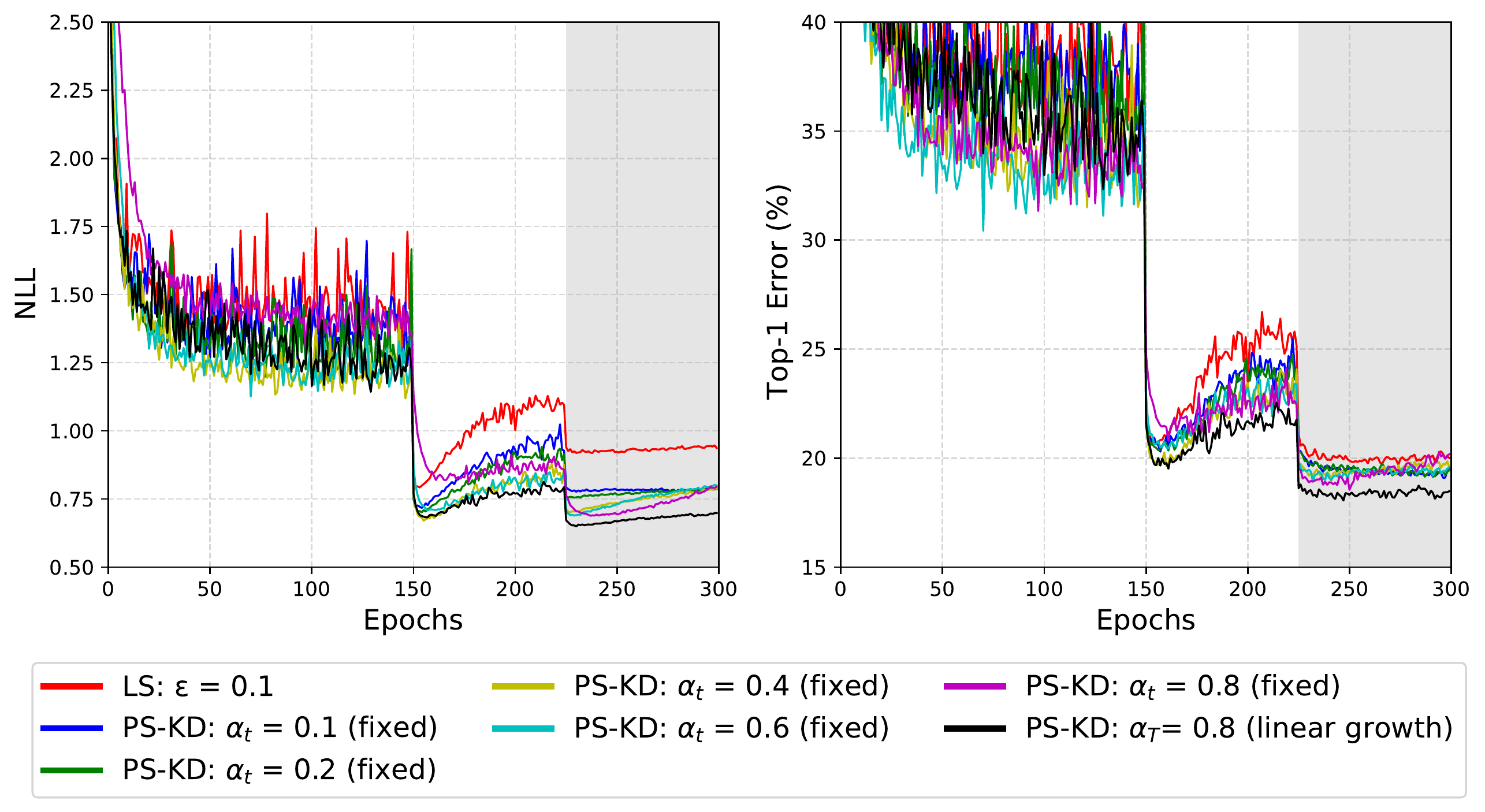}
  \caption{NLL (left) and top-1 error (right) curves on CIFAR-100 with different $\alpha_t$ values for DenseNet-121. Linear growth with $\alpha_T=0.8$ achieves the lowest NLL and top-1 error.}
  \vspace{-0pt}
  \label{figure:curve}
\end{figure}

\begin{figure*}[h]
  \centering
     \begin{subfigure}[b]{0.33\textwidth}
         \includegraphics[width=\textwidth]{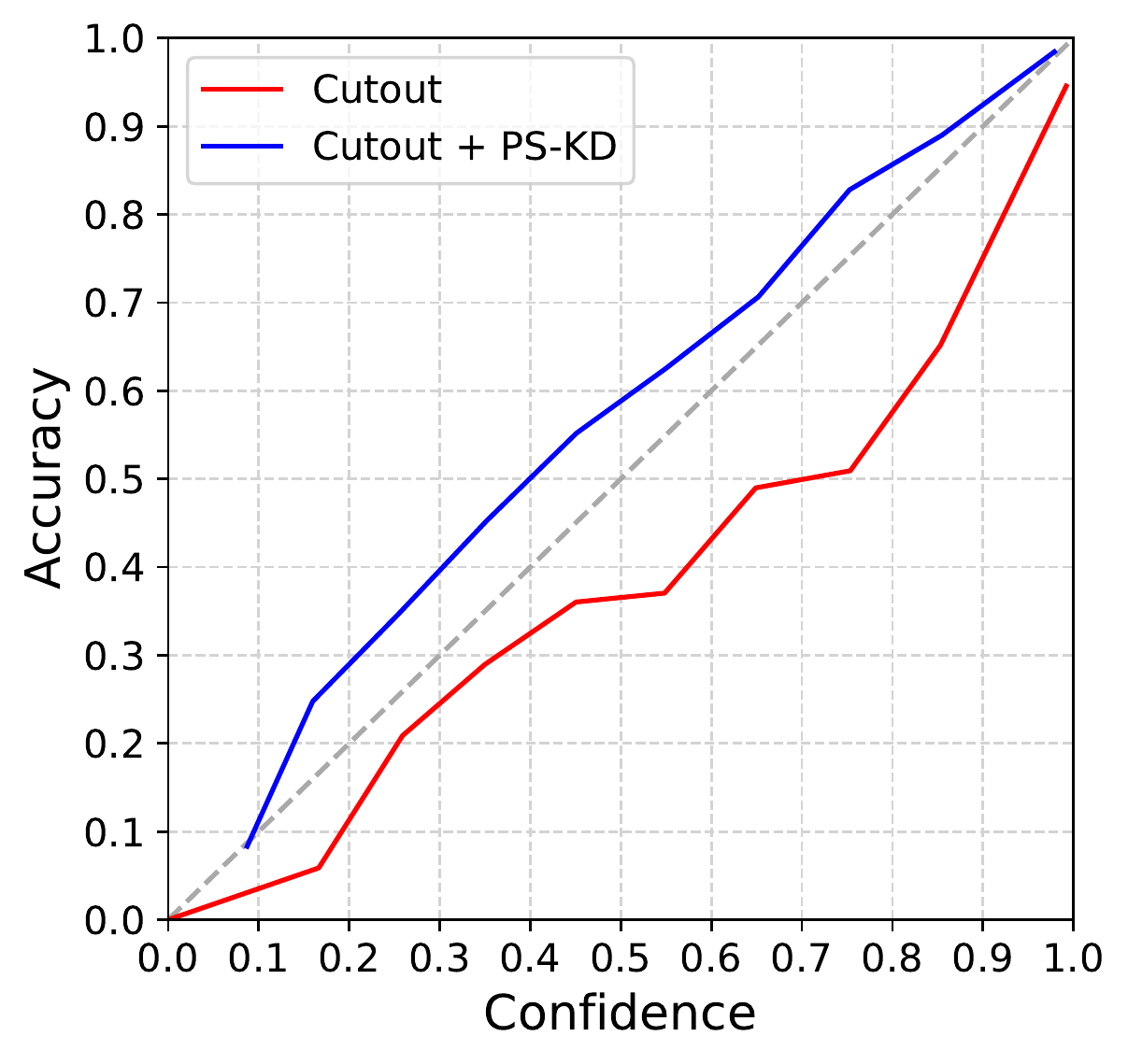}
         \caption{PyramidNet with Cutout}
     \end{subfigure}
     \begin{subfigure}[b]{0.33\textwidth}
         \includegraphics[width=\textwidth]{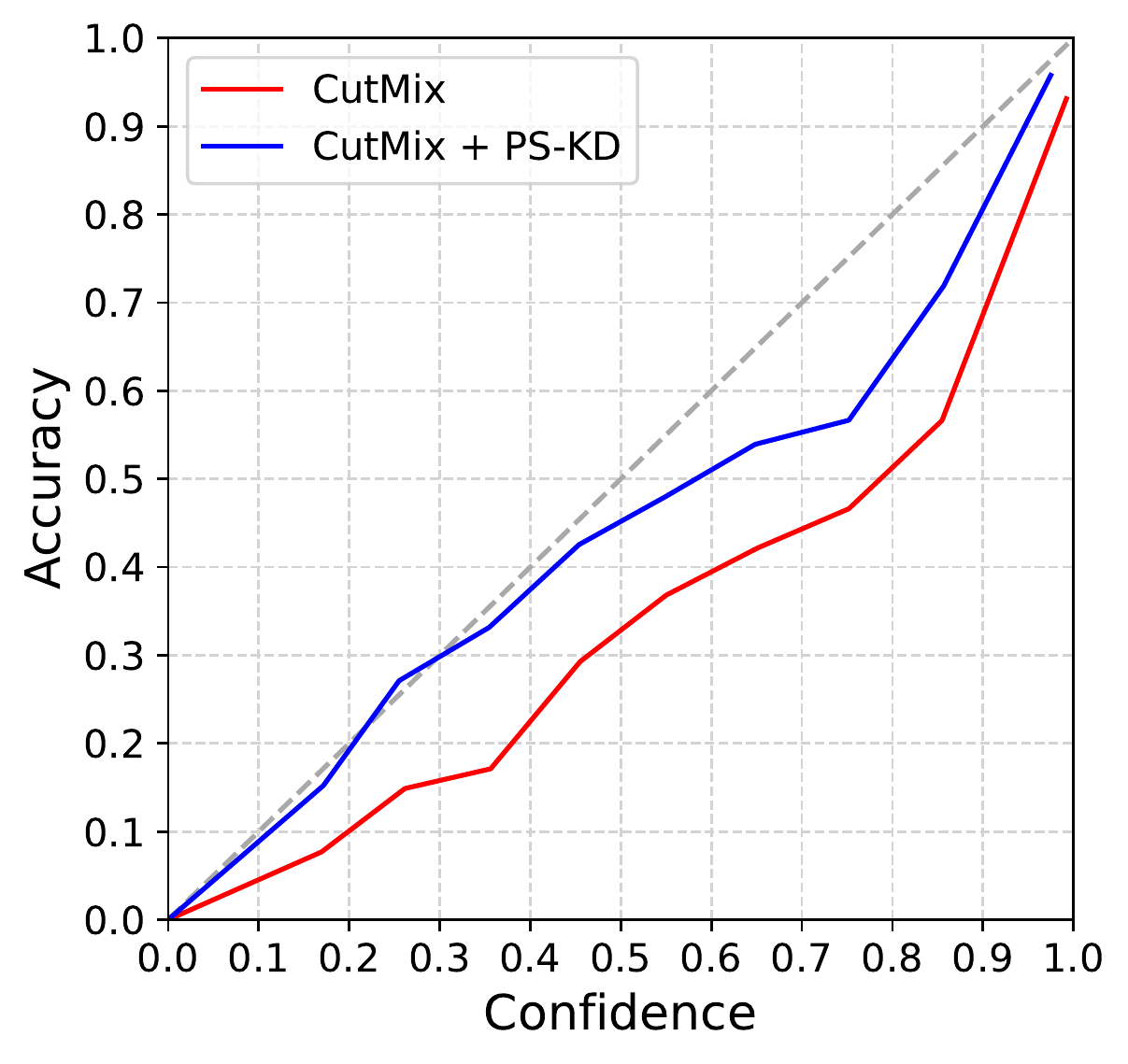}
         \caption{PyramidNet with CutMix}
     \end{subfigure}
     \begin{subfigure}[b]{0.33\textwidth}
         \includegraphics[width=\textwidth]{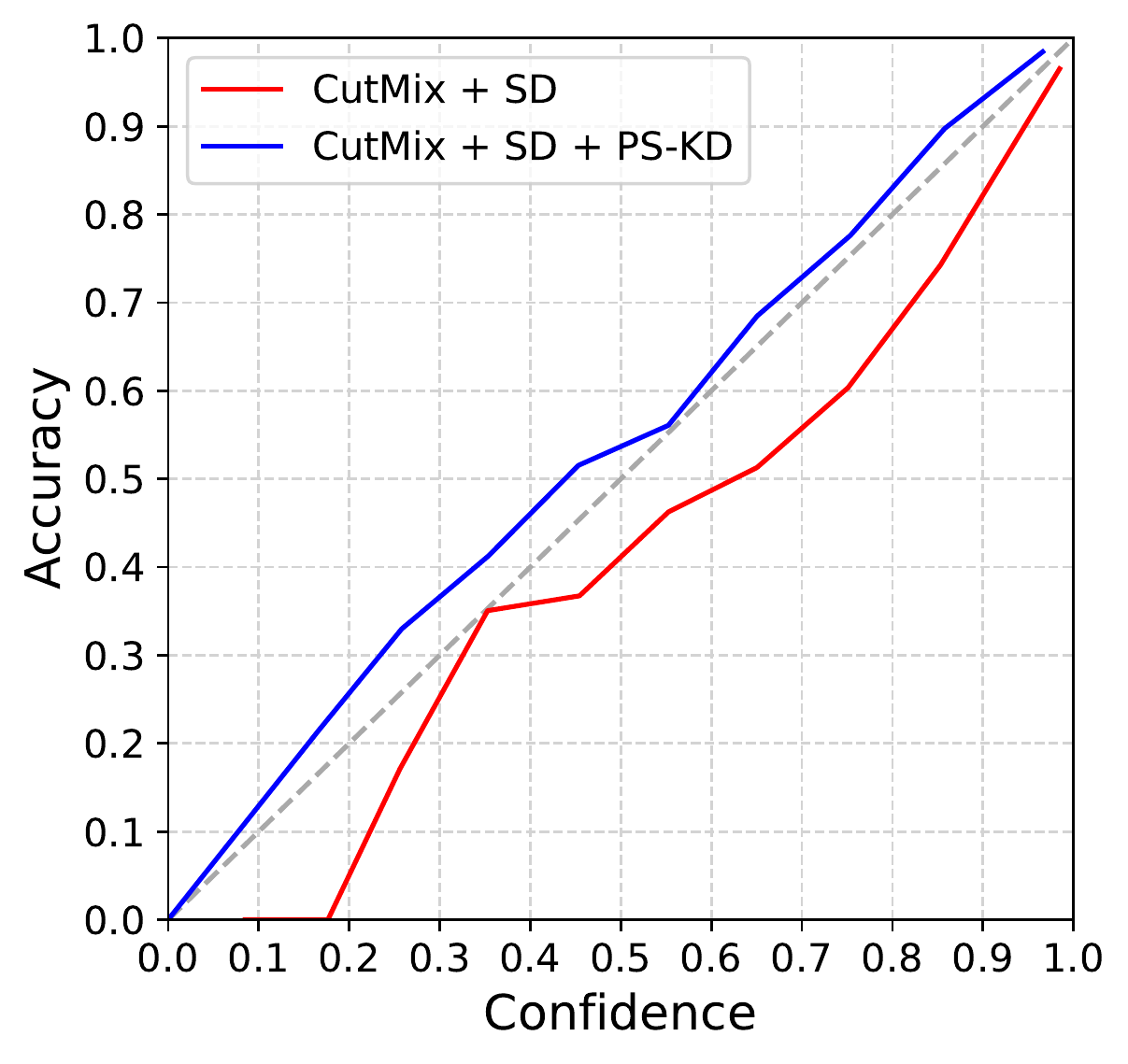}
         \caption{PyramidNet with CutMix + ShakeDrop}
     \end{subfigure}
        \vspace{-10pt}
        \caption{Calibration plots of advanced regularization methods on CIFAR-100 with PyramidNet. PS-KD provides additional benefits to existing methods in terms of calibration.}
        \vspace{-10pt}
        \label{fig:supple_reliability_of_combine_regularization on pyramid}
\end{figure*}

Additionally, to examine the effect of using past predictions to soften hard targets, we conduct experiments with a fixed value of $\alpha_t\in\{0.1,0.2,0.4,0.6,0.8\}$ so that the effect of adjusting $\alpha_t$ is excluded. From the curves of NLL and top-1 error in Fig.~\ref{figure:curve}, we observe that PS-KD with a fixed $\alpha_t=0.1$ shows lower NLL and top-1 error than LS with $\epsilon=0.1$ (refer to the shaded area on the curves), and the performances are improved as a fixed $\alpha_t$ increases. Therefore, it can be concluded that softening hard targets with predictions from the model itself is much better than just using a static softening operation like LS.
To further investigate the effect of adjusting $\alpha_t$, the curves from the linear growth strategy toward $\alpha_T=0.8$ are also depicted. Compared to the curves from the fixed $\alpha_t=0.8$, we conclude that the simplest approach, the linear growth, works surprisingly well for regularizing the model.
\newpage
\paragraph{Additional calibration plots}
Fig.~\ref{fig:supple_reliability_of_combine_regularization on pyramid} shows the calibration plots of existing regularization methods on CIFAR-100. From this figure, we can observe that the advanced regularization methods such as Cutout, CutMix, CutMix+SD benefit from PS-KD in terms of calibration.

\paragraph{Extension results for self-KD methods combined with advanced data augmentations}
As summarized in Table~\ref{table:additional_CIFAR-100 results}, we provide additional experimental results: [Cutout, CutMix, CutMix+SD] + LS, CS-KD, and TF-KD on CIFAR 100 with PyramidNet. The results show that PS-KD can be effectively combined with advanced regularization techniques.
\label{section:appendix_combine_result}
\begin{table}[h]
\small
\begin{center}
  \begin{tabular}{l|cc|ccc}
    \toprule
    \begin{tabular}[c]{@{}l@{}}\textbf{Model} \\ \textbf{+ Method}\end{tabular} & \begin{tabular}[c]{@{}c@{}}\textbf{Top-1} \\ \textbf{Err (\%)}\end{tabular} & \begin{tabular}[c]{@{}c@{}}\textbf{Top-5} \\ \textbf{Err (\%)}\end{tabular} &\textbf{NLL} &\begin{tabular}[c]{@{}c@{}}\textbf{ECE} \\ \textbf{(\%)}\end{tabular}&\begin{tabular}[c]{@{}c@{}}\textbf{AURC} \\ \textbf{($\times10^3$)}\end{tabular}\\
    \midrule
    PyramidNet & 16.80 & 3.69 & 0.73 & 8.04 & 36.95 \\
    + LS &{17.82} &{4.72} & {0.89} &{3.46} & {105.02}\\
    + CS-KD &{18.31} &{5.70} & {1.17} &{14.70} &{70.05}\\
    + TF-KD &{16.48} & {3.37} & {0.79} & {10.48} & {37.04}\\
    + PS-KD & \textbf{15.49} & \textbf{3.08} & \textbf{0.56} & \textbf{1.83} & \textbf{32.14}\\
    \cmidrule{1-6}
    + Cutout & 16.05 & 3.42 & 0.67 & 7.15 & 33.20\\
    + Cutout + LS & 17.15 & 4.38 & 0.82 & 4.65 & 82.61\\
    + Cutout + CS-KD & 18.20 & 5.25 & 1.06 & 13.78 & 66.69\\
    + Cutout + TF-KD & 16.29 & 3.18 & 0.74 & 9.77 & 35.78\\
    + Cutout + PS-KD & \textbf{14.82} & \textbf{2.86} & \textbf{0.54} & \textbf{3.69} & \textbf{29.77} \\
    \cmidrule{1-6}
    + CutMix & 15.62 & 3.38 & 0.68 & 8.16 & 34.60\\
    + CutMix + LS & 15.68 & 3.66 & 0.70 & \textbf{4.60} & 37.71\\
    + CutMix + CS-KD & 15.89 & 3.60 & 0.73 & 9.28 & 35.47\\
    + CutMix + TF-KD & 16.61 & 3.29 & 0.66 & 7.47 & 36.57\\
    + CutMix + PS-KD & \textbf{15.03} & \textbf{2.91} & \textbf{0.58} & 5.81 & \textbf{30.22} \\
    \cmidrule{1-6}
    + CutMix + SD & 14.07 & 2.38 & 0.51 & 3.96 & 28.65 \\
    + CutMix + SD + LS & 14.05 & 2.37 & 0.54 & \textbf{2.54} & 33.09 \\
    + CutMix + SD + CS-KD & 14.99 & 2.56 & 0.56 & 3.27 & 34.40 \\
    + CutMix + SD + TF-KD & 15.34 & 2.58 & 0.53 & 3.31 & 31.41 \\
    + CutMix + SD + PS-KD & \textbf{13.59} & \textbf{2.18} & \textbf{0.49} & 3.46 & \textbf{25.98} \\
    \bottomrule
  \end{tabular}
  \vspace{+8pt}
  \caption{Performance evaluation of self-KD methods with advanced data augmentation techniques. The values averaged over three runs are reported. The best result is shown in boldface.}
  \label{table:additional_CIFAR-100 results}
\end{center}
\end{table}

\newpage
\subsection{Experimental Results on ImageNet}
\paragraph{Random search results of the hyperparameters.}
To find out the optimal hyperparmeter of CS-KD, TF-KD and PS-KD, we perform a random search of hyperparameters over five trials with ResNet-152 for a fair comparison. We set the mini-batch size to 512, and the other training setting is set to the same as ImageNet experiments in the main manuscript. For CS-KD, we consider the range of the hyperparameters as follow: $\tau\in\{1, 2,\cdots, 20\}$ and $\lambda_{cls}\in\{0.1,0.5,1,\cdots, 4\}$. For TF-KD, we use TF-KD$_{reg}$ method which shows better performance on ImageNet in the original paper~\cite{TF-KD}. We consider the hyperparamters, the temperature $\tau\in\{20, 30, 40\}$, weight $\alpha\in\{0.1, 0.2,\cdots, 0.5\}$ and probability for the ground-truth class $a\in\{0.90,0.91,\cdots, 0.99\}$. For PS-KD, the range of $\alpha_T\in\{0.1, 0.2, \cdots, 1\}$ is used. The results are presented in Table~\ref{table:random 1}.
\label{section:appendix_RandomSearch}
\begin{table}[h]
\small
\begin{center}
 \begin{tabular}{l|cc|ccc}
    \toprule
    \begin{tabular}[c]{@{}l@{}}\textbf{Model} \\ \textbf{+ Method}\end{tabular} & \begin{tabular}[c]{@{}c@{}}\textbf{Top-1} \\ \textbf{Err (\%)}\end{tabular} & \begin{tabular}[c]{@{}c@{}}\textbf{Top-5} \\ \textbf{Err (\%)}\end{tabular} &\textbf{NLL} &\begin{tabular}[c]{@{}c@{}}\textbf{ECE} \\ \textbf{(\%)}\end{tabular}&\begin{tabular}[c]{@{}c@{}}\textbf{AURC} \\ \textbf{($\times10^3$)}\end{tabular}\\
    \midrule
    ResNet-152 & 21.95 & 6.16 & 0.89 & 5.08 & 61.64 \\
    + LS & 21.80 & 6.03 & 0.94 & 3.42 & 70.83\\
    \cmidrule{1-6}
    + CS-KD ($\tau=10, \lambda=4$) & 23.28 & 7.02 & 1.04 & 4.31 & 69.68\\
    + CS-KD ($\tau=20, \lambda=2$) & 22.30 & 6.46 & 0.95 & 4.92 & \textbf{54.13}\\
    + CS-KD ($\tau=1, \lambda=0.1$) & 21.68 & 6.04 & \textbf{0.85} & \textbf{1.46} & 61.09\\
    + CS-KD ($\tau=4, \lambda=0.5$) & \textbf{21.67} & \textbf{6.01} & 0.88 & 3.79 & 61.39\\
    + CS-KD ($\tau=10, \lambda=3$) & 22.43 & 6.55 & 0.98 & 5.45 & 65.99\\
    \cmidrule{1-6}
    + TF-KD ($\alpha=0.3,\tau=20,a=0.91$) & 22.72 & 6.49 & 0.92 & 4.69 & 65.30\\
    + TF-KD ($\alpha=0.1,\tau=40,a=0.95$) & \textbf{22.66} & \textbf{6.46} & \textbf{0.91} & \textbf{4.61} & \textbf{64.29}\\
    + TF-KD ($\alpha=0.2,\tau=40,a=0.97$) & 22.99 & 6.66 & 0.93 & 5.13 & 65.69\\
    + TF-KD ($\alpha=0.3,\tau=40,a=0.92$) & 22.82 & 6.61 & 0.92 & 4.72 & 64.79\\
    + TF-KD ($\alpha=0.1,\tau=30,a=0.92$) & 22.74 & 6.52 & 0.92 & 5.25 & 64.76\\
    \cmidrule{1-6}
    + PS-KD ($\alpha_T=0.9$) & 22.69 & 6.44 & 1.06 & 17.1 & 69.75\\
    + PS-KD ($\alpha_T=0.5$) & 21.67 & 5.92 & 0.88 & 7.33 & 63.19\\
    + PS-KD ($\alpha_T=0.1$) & 21.89 & 6.00 & 0.86 & 2.96 & 60.88\\
    + PS-KD ($\alpha_T=0.3$) & \textbf{21.51} & \textbf{5.86} & \textbf{0.84} & \textbf{1.85} & \textbf{60.61}\\
    + PS-KD ($\alpha_T=0.8$) & 22.40 & 6.40 & 1.00 & 13.65 & 68.00\\
    \bottomrule
 \end{tabular}
 \vspace{+8pt}
 \caption{Results over five trials of random search with ResNet-152. The best result for each method is shown in boldface.}
 \label{table:random 1}
\end{center}
\end{table}

\begin{figure*}[ht]
  \centering
     \begin{subfigure}[b]{0.33\textwidth}
         \includegraphics[width=\textwidth]{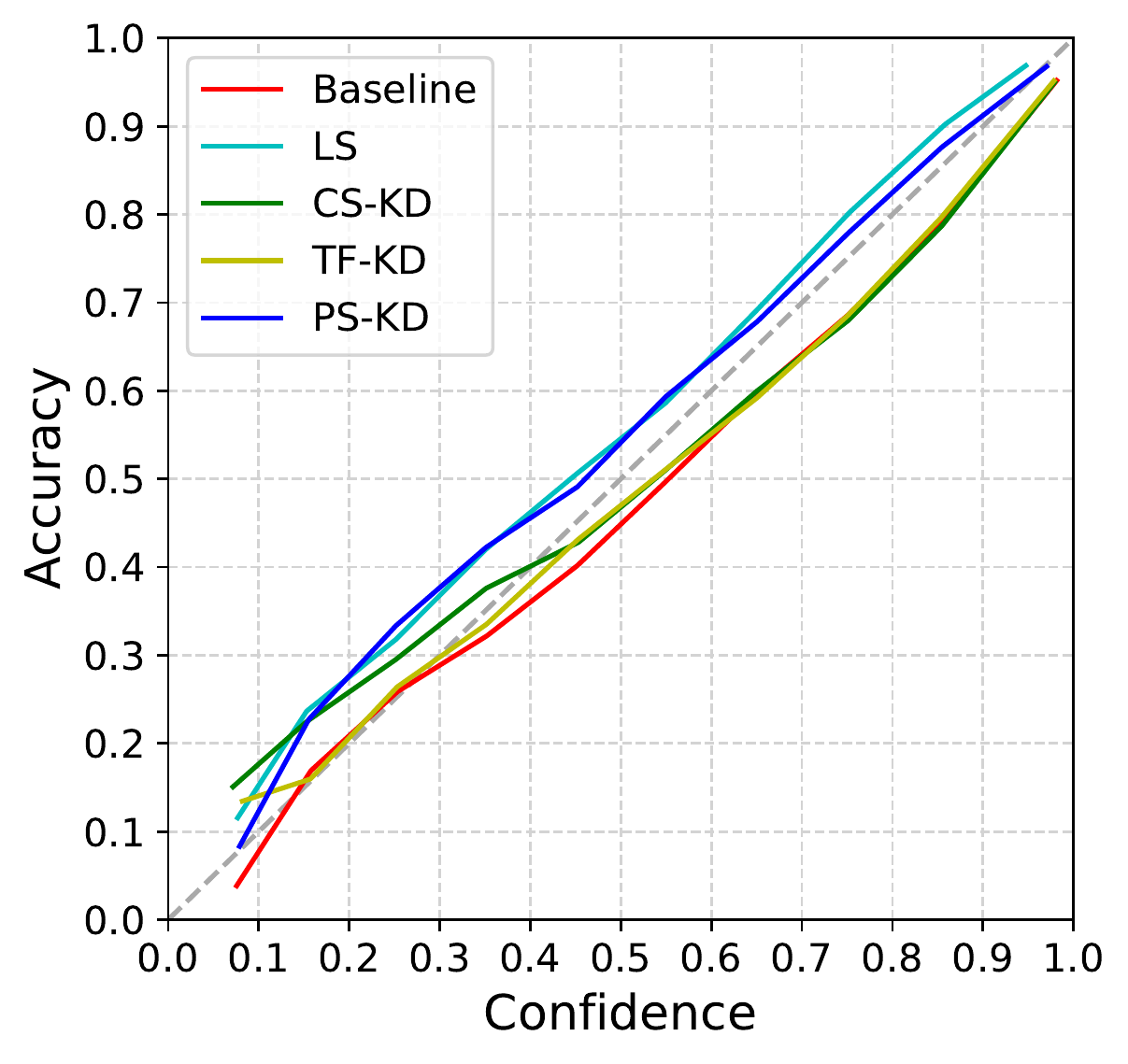}
         \caption{ResNet-152 with other methods}
     \end{subfigure}
     \begin{subfigure}[b]{0.33\textwidth}
         \includegraphics[width=\textwidth]{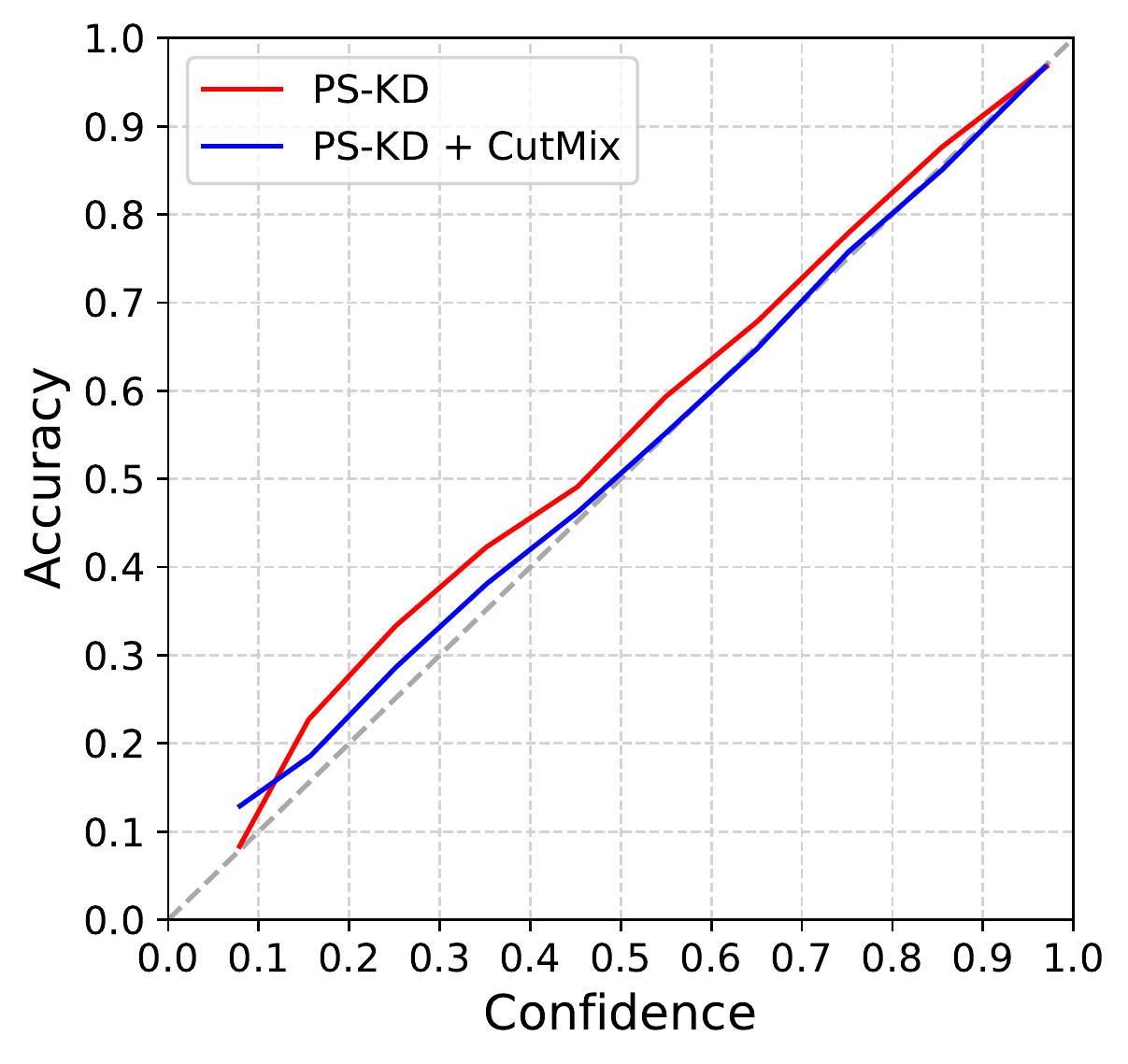}
         \caption{ResNet-152 with PS-KD + CutMix }
     \end{subfigure}
        \caption{Calibration plots on ImageNet with ResNet-152. (a) PS-KD shows slightly better performance compared to LS, CS-KD, and TF-KD. (b) PS-KD provides additional benefits to CutMix in terms of calibration.}
        \vspace{-8pt}
        \label{fig:supple_reliability_of_imagenet}
\end{figure*}

\paragraph{Additional calibration plots}
Fig.~\ref{fig:supple_reliability_of_imagenet} shows the calibration plots of comparison targets and CutMix. From this figure, we observe that PS-KD is better calibrated than other methods as well as improves calibration performance of the existing advanced regularization method, CutMix.
\newpage

\begin{figure*}[h]
    \centering
    \resizebox{\textwidth}{!}{%
    \begin{subfigure}{.4\textwidth}
        \centering
        \includegraphics[width=1.0\linewidth]{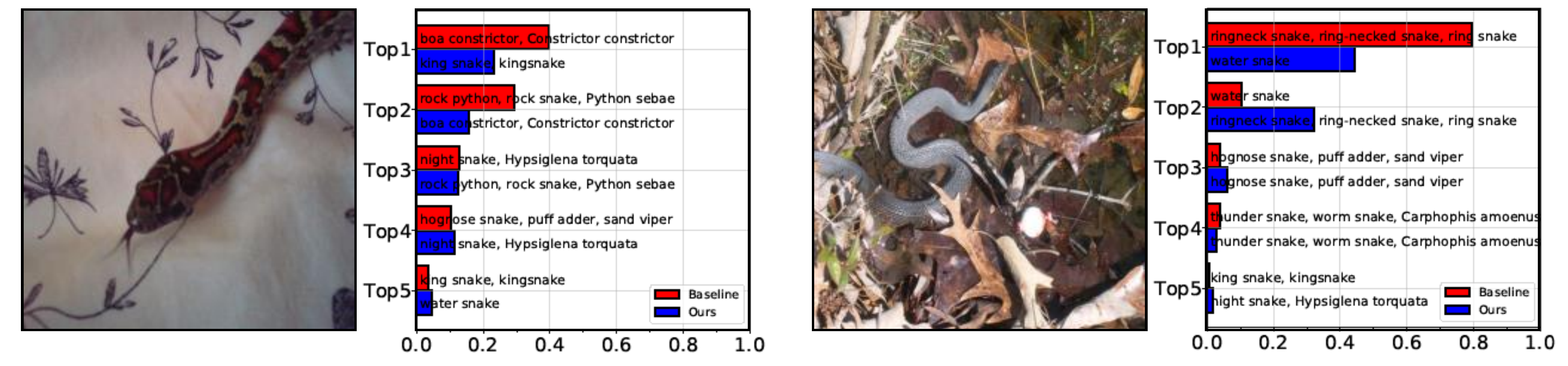}
    \end{subfigure}}
    \resizebox{\textwidth}{!}{%
    \begin{subfigure}{.4\textwidth}
        \centering
        \includegraphics[width=1.0\linewidth]{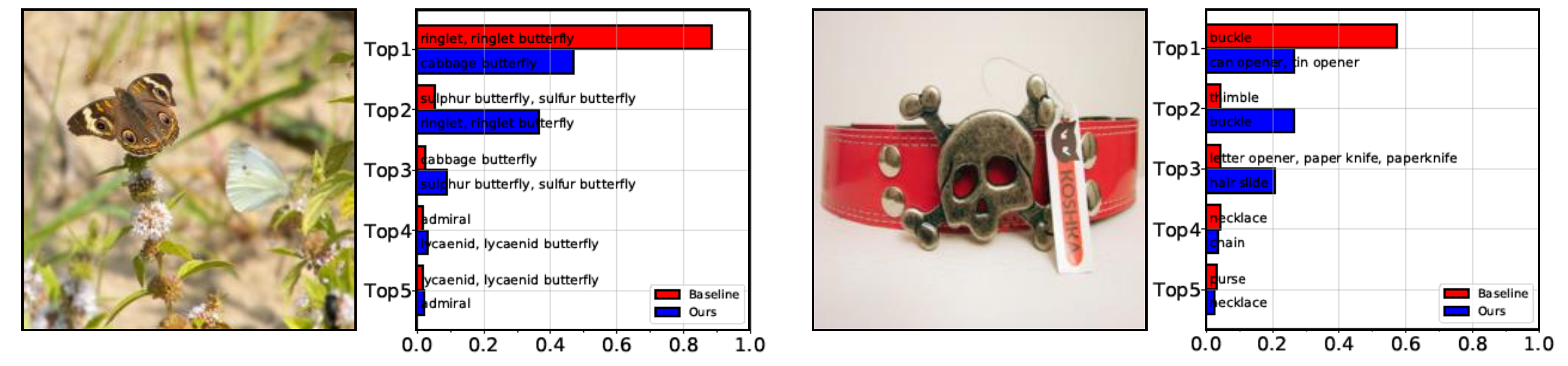}
    \end{subfigure}}
    \resizebox{\textwidth}{!}{%
    \begin{subfigure}{.4\textwidth}
        \centering
        \includegraphics[width=1.0\linewidth]{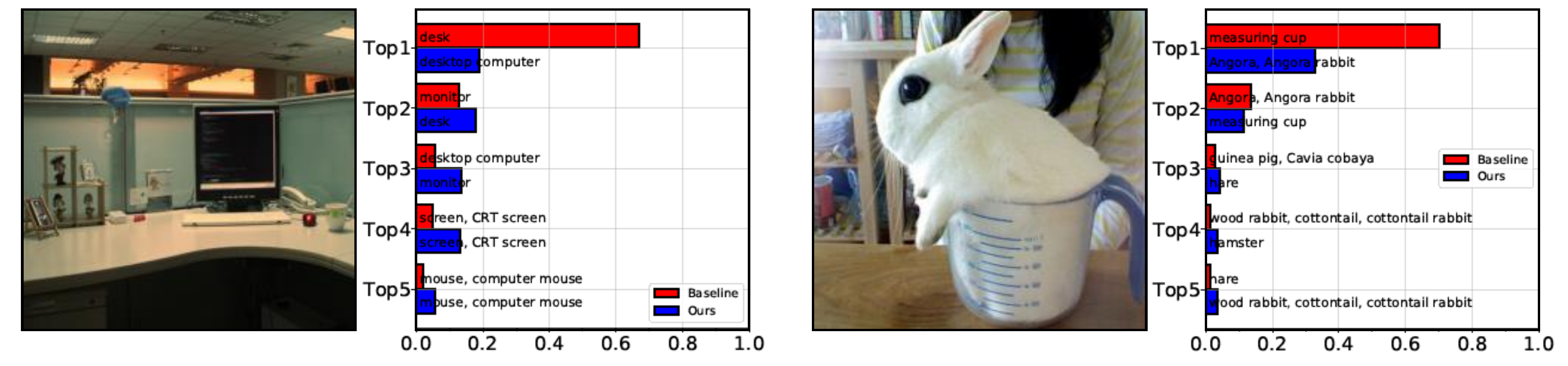}
    \end{subfigure}}
    \resizebox{\textwidth}{!}{%
    \begin{subfigure}{.4\textwidth}
        \centering
        \includegraphics[width=1.0\linewidth]{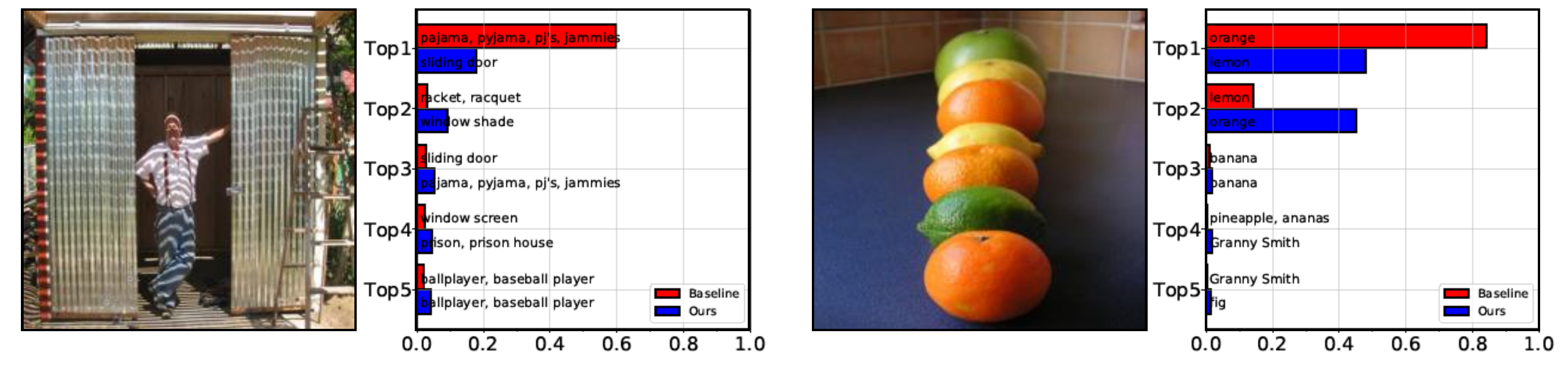}
    \end{subfigure}}
  \vspace{-5pt}
  \caption{Predicted probabilities for sample images from the baseline and PS-KD. From the top left, the ground-truth labels of these images are "king snake", "water snake", "cabbage butterfly", "buckle", "desk", "measuring cup", "sliding door" and "orange", respectively.}
  \vspace{-8pt}
  \label{figure:imgsmaples_supply}
\end{figure*}

\paragraph{Additional samples from ImageNet validation dataset.}
In Fig.~\ref{figure:imgsmaples_supply}, additional samples from ImageNet validation dataset and their predicted probabilities are presented. From these samples, we observe that PS-KD provides better outputs in the sense of human interpretation.

\newpage
\section{Object Detection}
Table~\ref{object_ap} shows the values of average precision (AP) over all classes. PS-KD shows higher AP values than the baseline and other methods (i.e., LS, CS-KD and TF-KD) for 10 classes out of 20 classes.

\begin{table*}[ht]
\centering
    \resizebox{\textwidth}{!}{%
    \begin{tabular}{l|cccccccccc|c}
    \toprule
    \multicolumn{1}{c}{Method} & \multicolumn{10}{c}{Average Precision} & \multicolumn{1}{c}{mAP}\\
    \midrule
    &Aeroplane & Bicycle & Bird & Boat & Bottle & Bus & Car & Cat & Chair & Cow & \multirow{16}{*}{\begin{tabular}[c]{@{}c@{}} 78.26\\ 78.44 \\ 78.33 \\ 78.28 \\ 79.50 \\ \textbf{79.72}\end{tabular}} \\
    \cmidrule{1-11}
    ResNet-152 & 78.87 & 84.70 & 79.64 & 72.60 & 63.45 & 86.68 & 87.38 & 88.08 & 63.27 & 83.29 &\\
    +LS& 81.52 & 84.97 & 79.53 & 69.58 & 63.71 & 83.88 & 87.25 & 87.49 & 64.39 & 85.15 &\\
    +CS-KD& 79.93 & 82.58 & 78.97 & 70.91 & \textbf{65.34} & 84.56 & 87.20 & 87.40 & 62.18 & 83.96 &\\
    +TF-KD& 79.97 & 85.98 & 78.57 & 70.79 & 61.45 & 85.96 & 87.69 & 87.86 & 61.24 & 85.00 &\\
    +PS-KD& 79.59 & 83.60 & \textbf{79.74} & 70.24 & 64.64 & \textbf{87.30} & \textbf{88.39} & 88.04 & \textbf{65.48} & \textbf{86.75} &\\
    +PS-KD + CutMix& \textbf{83.54} & \textbf{85.99} & 79.23 & \textbf{72.69} & 65.08 & 86.66 & 88.23 & \textbf{88.93} & 64.14 & \textbf{86.75} &\\
    \cmidrule{1-11}
    & \begin{tabular}[c]{@{}c@{}}Dining\\ Table\end{tabular} & Dog & Horse & \begin{tabular}[c]{@{}c@{}}Mortor\\ Bike\end{tabular} & Person & \begin{tabular}[c]{@{}c@{}}Potted\\ Plant\end{tabular} & Sheep & Sofa & Train & \begin{tabular}[c]{@{}c@{}}TV\\ Monitor\end{tabular} & \\
    \cmidrule{1-11}
    ResNet-152& 69.14 & 87.10 & 87.27 & 82.72 & 79.31 & 52.29 & 78.77 & 78.50 & 84.36 & 77.84 &\\
    +LS& 73.17 & \textbf{87.82} & 87.32 & 80.76 & 80.97 & 50.88 & 79.60 & 78.87 & 87.18 & 74.81 &\\
    +CS-KD& 72.47 & 85.17 & 87.41 & 82.08 & 81.32 & 53.43 & 82.04 & 76.97 & 85.40 & 77.33 &\\
    +TF-KD& 74.04 & 85.80 & 86.68 & 80.92 & 79.02 & 52.25 & 81.68 & 77.81 & 85.53 & 77.39 &\\
    +PS-KD& \textbf{77.98} & 87.68 & 87.55 & \textbf{85.07} & \textbf{81.42} & 53.15 & \textbf{82.50} & \textbf{79.78} & 83.64 & 77.45 &\\
    +PS-KD + CutMix& 71.98 & 87.10 & \textbf{87.73} & 84.01 & 81.34 & \textbf{56.05} & 78.07 & 79.76 & \textbf{87.92} & \textbf{79.21} &\\                  
    \bottomrule
    \end{tabular}}
    \vspace{+3pt}
    \caption{APs over all classes on PASCAL VOC 2007 testset. The best result for each class is in bold.}
    \label{object_ap}
\end{table*}

\section{Machine Translation}
\subsection{Evaluation Metrics}
\paragraph{BLEU.} BLEU (Bilingual Evaluation Understudy) is an algorithm for numerically measuring the quality of machine translation from one natural language to another one. By using human translation as a reference, BLEU evaluates the quality of machine translation via two aspects. One is how many $n$-grams in the translated output of a model appears in the reference. If the more $n$-grams appear in both machine translation and human translation, the quality of machine translation is considered as better. We set $n$ to 4, which is generally used for the evaluation.
Another aspect of BLEU is the length of machine translated sentence. If we evaluate the performance by using only $n$-grams, very short sentence with only few words in the reference will have nearly a perfect score. To prevent this, an additional term comparing the length of machine translation and human translation is considered in the calculation of BLEU.

\subsection{Dataset}
\paragraph{Dataset.} We use IWSLT15 English to German (EN-DE) and German to English (DE-EN) dataset. It consists of 191K training sentence pairs\footnote{The dataset can be downloaded from \href{https://https://wit3.fbk.eu/2015-01}{https://https://wit3.fbk.eu/2015-01}.}, and 8,300 pairs of the training data are used for validation. We concatenate dev2010, dev2012, tst2010, tst2011, tst2012, tst2013 datasets for a test set.

\end{document}